\documentclass{article}
\usepackage{kr}

\usepackage{mathptmx}
\DeclareMathAlphabet{\mathcal}{OMS}{cmsy}{m}{n}

\usepackage{soul}
\usepackage{url}
\usepackage{hyperref}
\usepackage[small]{caption}
\usepackage{graphicx}
\usepackage{mathtools}
\usepackage{amsthm}
\usepackage{amssymb}
\usepackage{booktabs}
\usepackage{algorithm}
\usepackage{algorithmic}
\usepackage{multicol}
\usepackage{tikz-cd}
\usetikzlibrary{shapes.geometric, arrows.meta, positioning, calc, shapes.symbols, shapes.misc, decorations.markings, shadows}
\usepackage{booktabs, threeparttable, multirow}
\usepackage{wrapfig}

\usepackage{xcolor}

\definecolor{DodgerUniformBlue}{rgb}{0.0,0.353,0.612}
\newcommand{\define}[1]{\emph{\textcolor{DodgerUniformBlue}{#1}}}

\definecolor{paleyellow}{HTML}{FFEC7F}

\long\def\comment[#1]#2{\medskip\par\noindent\colorbox{paleyellow}{\llap{\footnotesize #1:\quad}%
    \parbox[t]{\textwidth}{\setlength{\parskip}{1ex plus 0.2ex minus 0.2ex}#2}}}

\def\void{}
\newcommand{\mcomment}[2][\void]{\relax}

\usepackage{xspace}
\usepackage{tabularx}
\newcolumntype{L}{>{$}l<{$}}
\usepackage{stmaryrd}
\newcommand{\sem}[1]{\llbracket #1\rrbracket}
\newcommand{\para}[1]{\mathop{\mathrm{#1}}}
\newcommand{\fun}[1]{\mathop{\mathrm{#1}}}
\newcommand{\AWPO}{\mathrm{AWPO}}
\newcommand{\WPO}{\mathrm{WPO}}

\newcommand{\expsym}{\mathop{\raisebox{0.5ex}{\footnotesize$\wedge$}}}

\newcommand{\sk}{\para{sk}}
\DeclareMathOperator{\var}{\mathit{var}}

\DeclareMathOperator{\simp}{\mathit{simp}}
\DeclareMathOperator{\calF}{{\cal F}}
\DeclareMathOperator{\calA}{{\cal A}}
\DeclareMathOperator{\calB}{{\cal B}}

\DeclareMathOperator{\calI}{{\cal I}}

\newcommand{\rwsystemname}[1]{\text{\textsc{#1}}\xspace}
\newcommand{\ARI}{\rwsystemname{Ari}}
\newcommand{\NORM}{\rwsystemname{Norm}}
\newcommand{\CANON}{\rwsystemname{Canon}}
\newcommand{\SIMP}{\rwsystemname{Simp}}
\newcommand{\CLEAN}{\rwsystemname{Clean}}
\newcommand{\kwote}{\text{$\para{quote}$}\xspace} 

\newtheorem{lemma}{Lemma}
\newtheorem{theorem}[lemma]{Theorem}

\theoremstyle{definition}

\newtheorem{example}[lemma]{Example}

\newtheorem{note}[lemma]{Note}

\author{%
Peter Baumgartner$^1$\and
Lachlan McGinness$^2$ \\
\affiliations
$^1$CSIRO/Data61 and Australian National University\\
$^2$ Australian National University and CSIRO/Data61\\
\emails
\url{peter.baumgartner@data61.csiro.au},
\url{lachlan.mcginness@anu.edu.au}
}

\title{The AlphaPhysics Term Rewriting System for\\[0.4ex]
Marking Algebraic Expressions in Physics Exams}

\begin{document}

\maketitle

\begin{abstract}
We present our method for automatically marking Physics exams. The marking problem consists in assessing typed student answers for correctness with  respect to a ground truth solution. This is a challenging problem that we seek to tackle using a combination of a computer algebra system, an SMT solver and a term rewriting system. A Large Language Model is used to interpret and remove errors from student responses and rewrite these in a machine readable format.
Once formalized and language-aligned, the next step then consists in applying automated reasoning techniques for assessing student solution correctness. We consider two methods of automated theorem proving: off-the-shelf SMT solving and term rewriting systems tailored for physics problems involving trigonometric expressions. The development of the term rewrite system and establishing termination and confluence properties was not trivial, and we describe it in some detail in the paper. We evaluate our system on a rich pool of over 1500 real-world student exam responses from the 2023 Australian Physics Olympiad. 
\end{abstract}


\section{Introduction}

Many teachers across Australia are `burning out' and leaving the profession due to excessive workload \cite{Windle2022Teachers}. As marking is one of the largest contributions to teacher workload, teachers are seeking AI marking solutions to make their workload more sustainable \cite{Ogg2024Brisbane}. Automated Essay Grading \cite{Ramesh2021Automated} and Automated Short Answer Grading \cite{Weegar2024Reducing} are longstanding areas of research. 

Until recently, there were no effective strategies to mark free-form physics problems, as they could contain diverse inputs including text, equations and diagrams. Developments in generative AI have changed this and recent works have evaluated the potential of Large Language Models (LLMs) in grading physics exams \cite{Kortemeyer2023Toward,Kortemeyer2024Grading,mok_using_2024,Chen2025Grading,mcginness_can_2025}. 
However, there are no guarantees of the correctness of LLM reasoning \cite{Subbarao2024LLMmodulo}.

We propose a new framework, called AlphaPhysics,
that 
uses a combination of LLMs and automated reasoning engines. AlphapPhysics uses the strong pattern recognition abilities of LLMs to translate student responses into a standardized format before applying more rigorous reasoning engines, specifically \define{Term Rewriting Systems} (\define{TRS}s)~\cite{dershowitz_chapter_1990,Baader1998Term}, to evaluate student responses. 

Our TRSs are tailored for automating marking tasks in the pre-calculus physics domain.
The underlying rule language features rule axioms for addition, multiplication, exponentiation and
trigonometric functions (sine, cosine).
The main reasoning task is simplification of algebraic expressions to
a concise and human-readable normal form by, for
example, collecting like terms and evaluating arithmetic expressions to numbers when possible.
Normalization acts as a semantics-preserving ``translation'' service, 
and the normal form of a student's expression can be compared to a solution to determine if
they are semantically equivalent. 

Automated reasoning for Physics equations is challenging. Any (sound)
axiomatization of our set of trigonometric and arithmetic operators must
necessarily be incomplete. Specifically, applying term rewriting techniques faces
challenges as it requires reasoning on infinite domains with built-in (arithmetic)
operators and in presence of commutativity and other axioms. 
From that point of view, our wider research goal is to explore the potential of term
rewriting for automated marking in the Physics (and related) domain.

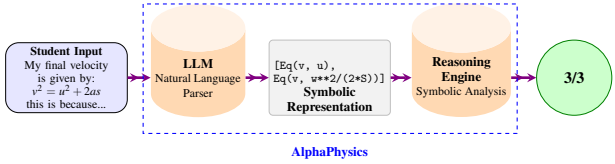
\begin{figure}
\centering
    \scalebox{0.5}{
    \begin{tikzpicture}[
        scale=1.0,
        every node/.style={transform shape},
        note/.style={
            draw=gray!30,
            fill=gray!5,
            rounded corners=5pt,
            text width=3cm,
            minimum height=2cm,
            align=center,
            font=\small
        }
    ]
    \tikzset{
        thought bubble/.style={
            rectangle,
            draw,
            rounded corners=10pt,  
            fill=blue!10,
            text width=3cm,       
            minimum width=2cm,    
            minimum height=1.2cm, 
            align=center,
            font=\footnotesize    
        },
        processor/.style={
            cylinder,
            cylinder uses custom fill,
            cylinder body fill=orange!30,
            cylinder end fill=orange!10,
            shape border rotate=90,
            aspect=0.5,
            minimum width=2.5cm,
            minimum height=3cm,
            align=center
        },
        code block/.style={
            rectangle,
            draw=gray!50,
            fill=black!5,
            rounded corners=3pt,
            minimum width=3cm,
            minimum height=2cm,
            align=left,
            font=\ttfamily\small
        },
        score display/.style={
            circle,
            draw=green!50!black,
            fill=green!20,
            minimum size=2cm,
            font=\Large\bfseries
        }
    }
    
    \node[thought bubble] (input) at (0,0) {
        \textbf{Student Input}\\
        My final velocity\\
        is given by:\\
        $v^2 = u^2 + 2as$\\
        this is because...
    };
    
    \node[processor] (llm) at (3.5,0) {
        \textbf{LLM}\\
        \small{Natural Language}\\
        \small{Parser}
    };
    
    \node[code block] (parsed) at (7,0) {
        [Eq(v, u),\\
         Eq(v, w**2/(2*S))]\\
    };
    
    \node[processor] (engine) at (10.5,0) {
        \textbf{Reasoning}\\
        \textbf{Engine}\\
        \small{Symbolic Analysis}
    };
    
    \node[score display] (score) at (13.5,0) {
        3/3
    };
    
    \foreach \i/\j in {input/llm, llm/parsed, parsed/engine, engine/score} {
        \draw[
            -stealth,
            line width=2pt,
            red!50!blue,
            decoration={
                markings,
                mark=at position .5 with {\arrow{stealth}}
            },
            postaction={decorate}
        ] (\i) -- (\j);
    }

    \draw[blue, dashed, thick] 
    ($(llm.north west)+(-0.2,0.7)$) rectangle 
    ($(engine.south east)+(0.6,-0.6)$);


    \node[blue, font=\bfseries] at ($(parsed.south)+(0,-1)$) {AlphaPhysics};
    \node[align=center, font=\bfseries] at ($(parsed.south)+(0,0.4)$) (caption) {Symbolic\\ Representation};
    \end{tikzpicture}
    }
    \caption{\label{fig:alphaphysics-pipeline} In the \textit{AlphaPhysics} pipeline a student response is parsed by an LLM before a symbolic reasoning engine determines the corresponding grade. 
    \vspace{-0.5cm}}
\end{figure}

\paragraph{Main contributions.}
In this paper we describe the current state of our \textit{AlphaPhysics} framework (See Figure \ref{fig:alphaphysics-pipeline}). 
This includes both the LLM and the theorem proving components, with a focus on the latter.
The theorem proving component consists of four TRS, chained for normalizing student equations.
We introduce conceptual contributions to term rewriting for verifying termination and
confluence of our TRS language. 
We show our TRS is terminating, 
but  not confluent, hence incomplete wrt.\ the equational theory induced by the rewrite rules.
We implemented a normalization procedure and evaluated our TRS on a rich pool of over 1500
real-world student responses. We compare our TRS to the Z3 automated theorem
prover~\cite{gurfinkel_arithmetic_2024}.  
Our experimental evaluation shows that our systems are ``complete enough'' for all our examples.






\paragraph{Related work in Term Rewriting.}
Term rewriting with built-in operations and numeric constraints 
has a long tradition \cite{goos_abstract_1989,goos_operational_1994,kop_term_2013}.
Commutativity rules like $x+y \to y+x$ require special attention to guarantee termination of
rewriting. They can be dealt in the \emph{ordered completion} TRS
framework~\cite{Martin1990Ordered}. However, the combination of these topics has not received
much attention.

As an alternative to term rewriting, one could consider first-order logic automated theorem proving (ATP) over built-in domains. 
The hierarchic superposition calculus~\cite{baumgartner_hierarchic_2013}, for instance, features search space restrictions by means of term ordering constraints. 
ATPs are geared for refutational completeness (not normalization) and are harder to control than TRSs. While the termination and confluence analysis for a TRS is expected to be done once and forall as an offline step, superposition-based ATPs generate formulas for restoring confluence by deriving formulas during proof search. This involves inferences among the axioms, which typically is not (and need not be) finitely bounded.

The rest of this paper is structured as follows. Section \ref{sec:LLM} describes the use of LLMs for pre-processing student equations.
In Section \ref{sec:formal-framework}, we introduce the architecture of our TRS. Section \ref{sec:Dataset} introduces the specific marking problem that we are trying to solve. Section \ref{sec:results} shows the results of experiments comparing our TRS to an SMT solver. Section \ref{sec:limitations} discusses the limitations of our system and how it could be extended to mark a wider variety of physics problems.

\section{Large Language Model Pre-processing}
\label{sec:LLM}

AlphaPhysics is designed to meet the Australian AI Ethics Framework which include environmental well-being, transparency and privacy protection \cite{AustraliaAIEthics2024}. To address these requirements we use local open source LLMs to extract features, such as equations, from student responses so they can be parsed by our TRS. This approach maintains student privacy by keeping data on a local system. Our proposed framework supports transparency as errors in the LLM feature extraction can be easily seen, contested and corrected by a student or teacher. The Alpha Physics approach makes teachers aware of the compute that they are using by running local models, exposing the normally hidden electricity usage and environmental cost of their usage. 

The local LLM is given the task of extracting student equations from typed student responses and converting these into a standardised format. The student may make syntax errors or use incorrect or undefined variables. The LLM is expected to compensate for these errors in a way similar to a human marker in order to extract what the student intended to write. For example a student may write:\\ 
\centerline{m1 x v0 + m2 = m1 x v1 + m2 x v2} \\
\noindent which the LLM converts to correct SymPy syntax: \\
\centerline{Eq(m\_1*v\_0 + m\_2, m\_1*v\_1 + m\_2*v\_2)} 

A previous study \cite{mcginness_can_2025} tested LLMs' capabilities to extract equations using the typed student responses to the 2023 Australian Physics Olympiad. A 14 billion parameter model, Phi 4 \cite{abdinPhi4TechnicalReport2024}, was able to complete the task of translating student equations with 73\% accuracy by using an `LLM-Modulo' \cite{Subbarao2024LLMmodulo} 
prompting technique where Z3 is used as a syntax checker. 
If the equation provided by the LLM fails the Z3 check, then the LLM is prompted again with feedback on its previous response to repair the equations. 

This paper focuses on the application of the \textit{AlphaPhysics} TRS to the ground truth equations for each student response. The next section defines our TRS tailored for pre-calculus physics problems.

\section{Term Rewriting Framework}
\label{sec:formal-framework}


\paragraph{Preliminaries.}
We assume standard notions of first-order logic theorem proving and term rewriting, see~\cite{Harrison2009Handbook,Baader1998Term}.
A \define{signature} $\Sigma$ is a collection of function and predicate symbols of given
fixed and finite arities. 0-ary function symbols are also called constants.
In this paper we fix ``arithmetic'' signatures $\Sigma$ to comprise of the binary function
symbols $+$, $\times$, $\expsym$ (exponentiation), $/$ (division), unary function symbols
$\sin$, $\cos$, a unary function symbol $\kwote$ explained below, the constant $\pi$,  
and all integer and finite decimal concrete number constants, e.g., $-1$, $3.141$. 

Physics equations often contain universally quantified variables $\para x, \para y, \para
z, \para \theta$ etc. In order not to confuse them with the meta-variables of logic, 
we consider extensions of $\Sigma$ with finitely many (Skolem) constants $\Pi$ representing these
variables and call them \define{parameters}.
We write $\Sigma_\Pi$ for the extended signature. 

Let $\Sigma_\Pi(V)$ denote the extension of $\Sigma_\Pi$ by a denumerable set of variables. Unless noted
otherwise, we assume the signature $\Sigma_\Pi(V)$, for some $\Pi$ left unspecified. Terms are
defined as usual, but we use infix notation and parentheses for writing terms made with
arithmetic function symbols. We write just $c$ instead of $c()$ for constants $c$.
A \define{parameter-free term} is a term without occurrences of parameters.
We write $\var(t)$ for the set of variables occurring in a term $t$.
A \define{ground term} is a term $t$ with $\var(t) = \emptyset$.
We use notions of \define{substitutions}, \define{instance}, \define{ground instance},
\define{matching} and \define{unification} in a standard way. The notation
$t_p[u]$ means that $t$ has a, not necessarily proper, subterm $u$ at position $p$. The
position index $p$ is left away if clear from the context or not important.
We use the letters $\gamma$, $\delta$ and $\sigma$ for substitutions, and $s$ and $t$ for terms. A \define{ground
substitution $\gamma$ for $t$} is a substitution such that $t\gamma$ is ground. We often
assume $t$ is clear from the context and leave ``for $t$'' away.

As for semantics, let $\calA_\pi$ be a $\Sigma_\Pi$-Algebra with the reals as carrier set that maps
all arithmetic function symbols in $\Sigma$ to the expected functions over the reals, and maps
every parameter in $\Pi$ to some real number as per the \define{parameter mapping $\pi$}.  An
\define{assignment $\alpha$} is a mapping from the variables to the reals. 
We write
$\calA_\pi(\alpha)$ for the usual interpretation function on terms,
where variables are interpreted according to $\alpha$. For ground terms we can
unambiguously write $\calA_\pi(t)$ instead of $\calA_\pi(\alpha)(t)$.

\paragraph{Quoted terms and simplification.}
The signature includes the distinguished unary function symbol ``\kwote''. For better readability we
write a \define{\kwote-term} $\kwote(t)$ as $\sem{t}$ and call $t$ the
\define{quoted term}.
A term is \define{\kwote-free} if it is not a quoted term and does not contain a \kwote-term.


Quote-terms are our mechanism for building-in arithmetic on numbers. 
Ground quoted terms are simplified to a number constant (like $\sem{2}$ for $\sem{1+1}$), and non-ground quoted terms can be replaced by a semantically equivalent terms (like $\sem{2\times x + 1}$ for $\sem{x+x+1}$). 
Formally, we define a
\define{simplifier}\footnote{A simplifier plays a similar role as a ``canonizer'' in~\cite{shostakDecidingCombinationsTheories1984}.}
as a total function  $\simp$ on \kwote-free and
parameter-free terms such that
\begin{itemize}
\item[(i)] if $\simp(t) = s$ then, for all assignments $\alpha$, $\calA(\alpha)(t) = \calA(\alpha)(s)$, 
\item[(ii)] for every number $n$, $\simp(n) = n$, 
\item[(iii)] if $t$ is ground then $\simp(t)$ is a number. \end{itemize}
Condition (i) is a soundness 
requirement and forbids $\simp$ to equate two semantically different terms (``no confusion''). 
Condition (iii) is needed for completeness of theorem proving over ground terms by rewriting.
We do not need and cannot enforce a more general completeness requirement.
We only need $\simp$ to be ``complete enough'' for non-ground terms so that 
critical pairs can be joined. 


We extend $\simp$ homomorphically to all terms as follows:
\begin{equation*}
\define{$t{\downarrow_{\simp}}$} =
\begin{cases*}
  t & if $t \in V$ or $t \in \Pi$ \\
  \sem{\simp(u)} & if $t = \sem{u}$ \\
  f(t_1{\downarrow_{\simp}},\ldots,t_n{\downarrow_{\simp}}) & if $t = f(t_1,\ldots,t_n)$
\end{cases*}
\end{equation*}
We say that $t$ is \define{(fully) simplified} iff $t = t{\downarrow_{\simp}}$


By design of our rewrite systems, in all derivations from ground terms, all \kwote-terms in all rule instances are always parameter-free, ground and quote-free, hence can be simplified to a number. Non-ground \kwote-terms are needed only for confluence and termination analysis.

The vast majority of the literature on term-rewriting over built-in domains uses rules with constraints. What we achieve with a \kwote-term $\sem{t}$ can be expressed by replacing it in the rule with a \emph{sorted} variable $x$ and adding the constraint $x = t$. 
See \cite{kop_term_2013} for a discussion and overview of rules with constraints. 
Notice that our approach makes explicit typing  unnecessary as quoting achieves the same. We found that this simple approach works well for our use case.

\paragraph{Constrained rewrite rules and normal forms.}
A \define{(constrained rewrite) rule
  $\rho$} is of the form $l \to r \mid C$ where $l$ and $r$ are terms such that $\var(r) \subseteq
\var(l)$, and the \define{constraint $C$} is a finite set of formulas whose free
variables are contained in $\var(l)$. Notice that $\var(\rho)$, the set of (free) variables occurring in $\rho$, is just $\var(l)$.
If $C = \emptyset$ we just write $l \to r$.
We assume the signature of the constraint language contains conjunction, so that
constraints can be taken as the conjunction of their elements. We model evaluation of constraints by assuming a
\define{satisfaction relation $\calI$} on constraints. 
 We write $\calI \models C$ instead of
$C \in \calI$. Notice that $\calI \models \emptyset$, as expected. We require that constraint satisfaction is \define{stable under
  substitution}: if $\calI \models C$ then $\calI \models C\delta$.\footnote{For soundness reasons; the free variables are used in
  a universal quantification context, so instances better be satisfied, too.}
If $\calI \models C\delta$ we call the resulting unconstrained rule $l\delta \Rightarrow r\delta$ an \define{ordinary instance (of $\rho$) (via
  $\delta$)}. (We choose $l\delta \Rightarrow r\delta $ as $l\delta \to r\delta$ is our convention for a rule with an empty constraint.)

The constraints $C$ can
be purely operational in rules for pre-processing terms (in the \NORM TRS, and stability is not an issue) or weighted path ordering  (WPO) constraints (in the \CANON and \SIMP TRSs), see
Section~\ref{sec:theorem-proving-by-normalization}.

We say that $\rho$ is \define{sound} iff $\calA_\pi(\alpha)(s) = \calA_\pi(\alpha)(t)$ holds
for every ordinary instance $s \Rightarrow t$ of $\rho$, every parameter mapping $\pi$ and every assignment $\alpha$.
All our rules are sound.

We say that a variable $x \in \var(\rho)$ is \define{background} (\define{foreground}) if it
occurs inside (outside) some \kwote-term in $l$ or $r$.
We say that the rule $\rho$ is \define{admissible} if all of the following hold:
  \begin{itemize}
  \item[(i)] No variable in $\var(\rho)$ is both background and foreground.
    \item[(ii)] For every $\sem{t}$ occurring in $l$, $t$ is either a number or a (background) variable.
    \item[(iii)] For every $\sem{t}$ occurring in $r$ or $C$, $t$ is parameter-free and quote-free.
\end{itemize}

Admissibility has a similar effect as working with a hierarchical sort

Conditions (i) and (iii) together have a similar effect as working with a hierarchical sort
structure where the \kwote-terms belong to the background sort, and all other terms are
foreground or background. Condition (ii) makes sure that matching quoted terms to quoted
terms is purely syntactic. For example, the rule $-\sem{x} \to \sem{-1\times x}$ is admissible but
$\sem{-1\times-1\times x} \to \sem{x}$ is not. Condition (iii) prevents nesting of quote terms. 

\paragraph{Rewriting and normal forms.}
Let $\rho = (l \to r \mid C)$ be an admissible rule.  We say that \define{$s$ is obtained from
  $t$ by (one-step) rewriting (with simplification)} and write $t \to_\rho s$ if
$t = t_p[u]$ for some non-variable term $u$ and position $p$, $u = l\delta$ for some substitution $\delta$,
$l\delta \Rightarrow r\delta$, and $s = t_p[(r\delta){\downarrow_{\simp}}]$.
A rewriting step is \define{ground} iff $t$ is ground (and hence
$s$ is ground, too).

For example, $(\sem{1} + \sem{2}) + \para a \to_\rho  \sem{3} + \para a$ if $\rho = (\sem{a} + \sem{b} \to  \sem{a+b})$.
All our rules of interest and all instances built will always be
admissible as a consequence of how they are computed.

A \define{rewrite system $R$} is a finite set of rewrite rules. We define the \define{$R$-rewrite
relation $\to_R$} as $t \to_R s$ iff $t \to_\rho s$ for some $\rho \in R$.
Let $\to_R^*$ be the transitive-reflexive closure of $\to_R$. 
We say that \define{$s$ is an $R$-normal form of $t$} iff
$t \to_R^* s$ but $s \nrightarrow_R s'$ for any $s'$. 
We define the \define{$R$-normal form relation ${\downarrow_R}$} as 
$t \mathop{\downarrow_R}  s$ iff $s$ is an $R$-normal form of $t$.

\subsection{Proving Physics Equations by Normalization}
\label{sec:theorem-proving-by-normalization}
\emph{Theorem proving}, the validity problem for $\Sigma(V)$-equations $\forall(s = t)$,
is phrased in our setting as ``does $\calA_\pi(s_{\sk})
= \calA_\pi(t_{\sk})$ hold for all parameter mappings $\pi$?'', where $s_{\sk}$ and
$t_{\sk}$ are $\Sigma_\Pi$-terms obtained from $s$ and $t$ by uniquely replacing every variable by
a parameter from $\Pi$.
This problem is, of course, not solvable in general. Our TRS method approximates solving it
in an incomplete but sound way by combining four rewrite systems into one procedure for normalization.
If the normalized versions of $s$ and $t$ are syntactically equal then the answer is ``yes'', otherwise ``unknown''.

The four rewrite systems are \NORM, \CANON, \SIMP and \CLEAN,
collectively called the \define{ARI rewrite systems}. For normalization, the systems are
chained, each exhaustively applying rewrite rules on the result of the previous one.
More formally, we define the \define{ARI normalization} relation as
$\mathop{\to_\ARI} = {\downarrow_\NORM} \circ {\downarrow_\CANON}  \circ {\downarrow_\SIMP} \circ {\downarrow_\CLEAN}$.
The composition operator, $\circ$, stands for application from left to right.
We say that a term $u$ is an \define{\ARI-normal form of $s$} iff $\text{norm}(s)  \to_\ARI u$. 
We say that $s$ and $t$ are \define{algebraically equal}, written as \define{$s \approx t$}, if
$s$ and $t$ have a common \ARI-normal form $u$. If the set of rewrite rules is complete then equations which can be converted from one to the other by standard algebraic operations should have the same normal form. In this case we say that the normal form is unique.

Let us explain the design and intention of the ARI rewrite systems and how they work together.
The \NORM system, see Table \ref{tab:normalize-reformatted},
implements several conceptually simple ``preprocessing'' operations. They are triggered by
decorating a target term $s$ as $\text{norm}(s)$.
\begin{table}[h]
    \centering
    \caption{The \NORM rewrite system. The rule conditions in the rightmost column are
      in Python syntax.}
    \label{tab:normalize-reformatted}
    \resizebox{\columnwidth}{!}{%
    \begin{tabular}{l l c l l}
        \hline
        N1.1 & $\text{norm}(\sem{x})$ & $\rightarrow$ & $\sem{x}$ &  \\
    N1.2 & $\text{norm}(x)$ & $\rightarrow$ & $\sem{x}$ & $\mathrm{is(\mathit{x}, \text{number})}$ \\
    N1.3 & $\text{norm}(x)$ & $\rightarrow$ & $\sem{1} \times (\mathit{x} \expsym \sem{1})$ &
                                                                                    $\mathrm{is(\mathit{x}, \text{parameter})}$ \\
    \hline
    N2.1 & $\text{norm}(\sin(n \times y))$ & $\rightarrow$ & $\text{norm}(\fun{sin\_n}(n,y))$ & if $\mathrm{is(\mathit{n},\ int)}$  \\
     & & & & $\mathrm{ \ and\ \mathit{n} \geq 0}$ \\
     \hline
    N2.2 & $\text{norm}(\sin(x \times y))$ & $\rightarrow$ & $\sin(\text{norm}(x \times y))$ & if $\mathrm{not\ (is(\mathit{x},\ int)} $  \\
    & & & & $\mathrm{\ and\ \mathit{x} \geq 0)}$\\
    \hline
    N2.3 & $\text{norm}(\sin(x))$ & $\rightarrow$ & $\sin(\text{norm}(x))$ & if $\mathrm{not\ (is\_funterm(\mathit{x})\ }$ \\
    & & & & $\mathrm{ and\ \mathit{x}.fun\ ==\ \times)}$ \\
    \hline
    N2.4 & $\text{norm}(\cos(n \times y))$ & $\rightarrow$ & $\text{norm}(\fun{cos\_n}(n,y))$ & if $\mathrm{is(\mathit{n},\ int)\ } $ \\
    & & & & $\mathrm{ and\ \mathit{n} \geq 0}$\\
    \hline
    N2.5 & $\text{norm}(\cos(x \times y))$ & $\rightarrow$ & $\cos(\text{norm}(x \times y))$ & $\mathrm{not\ (is(\mathit{x},\ int)\ } $ \\
    & & & & $\mathrm{ and\ \mathit{x} \geq 0)}$ \\
    \hline
    N2.6  & $\text{norm}(\cos(x))$ & $\rightarrow$ & $\cos(\text{norm}(x))$ & $\mathrm{not\ (is\_funterm(\mathit{x})\ } $ \\
    & & & & $\mathrm{ and\ \mathit{x}.fun\ ==\ \times)}$ \\
    \hline
    N3.1 & $\text{norm}(\fun{sin\_n}(n,x))$ & $\rightarrow$ & $\fun{sin\_n}(\fun{to\_succ}(n),\text{norm}(x))$ &  \\
    N3.2 & $\text{norm}(\fun{cos\_n}(n,x))$ & $\rightarrow$ & $\fun{cos\_n}(\fun{to\_succ}(n),\text{norm}(x))$ &  \\
    \hline
    N4.1 & $\text{norm}(x + y)$ & $\rightarrow$ & $\text{norm}(x) + \text{norm}(y)$ &  \\
    N4.2 & $\text{norm}(x \times y)$ & $\rightarrow$ & $\text{norm}(x) \times \text{norm}(y)$ &  \\
    \hline
    N4.3 & $\text{norm}(x \expsym y)$ & $\rightarrow$ & $\text{norm}(x) \expsym \text{norm}(y)$ & $\mathrm{not\ (is(\mathit{y},\ int)\ } $  \\
    & & & & $\mathrm{ and\ \mathit{y}\ \geq\ 0)}$ \\
    \hline
    N5.1 & $\text{norm}(x \expsym n)$ & $\rightarrow$ & $\text{norm}(\fun{pwr\_n}(x,n))$ & $\mathrm{is(\mathit{n},\ int)\ } $ 
    $\mathrm{ and\ \mathit{n} \geq 0}$\\
    N5.2 & $\text{norm}(\fun{pwr\_n}(x,n))$ & $\rightarrow$ & $\fun{pwr\_n}(\text{norm}(x),\fun{to\_succ}(n))$ &  \\
    N5.3 & $\fun{to\_succ}(0)$ & $\rightarrow$ & $\sem{0}$ &  \\
    N5.4 & $\fun{to\_succ}(n)$ & $\rightarrow$ & $\fun{s}(\fun{to\_succ}(\mathrm{\mathit{n}-1}))$ & $\mathrm{\mathit{n}\ >\ 0}$ \\
    \hline
    N6.1 & $\text{norm}(x - y)$ & $\rightarrow$ & $\text{norm}(x + \fun{uminus}(y))$ &  \\
    N6.2 & $\text{norm}(\fun{uminus}(x))$ & $\rightarrow$ & $\sem{-1} \times \text{norm}(x)$ &  \\
    \hline
    N7.1 & $\text{norm}(x / y)$ & $\rightarrow$ & $\text{norm}(x) \times (\text{norm}(y) \expsym \sem{-1})$ &  \\
        \hline
    \end{tabular}}
\end{table}

\NORM expands exponentiation terms $x\expsym n$ where $n$ is an integer number into a
term $\mathrm{pwr\_n}(x, \mathrm{s}(\mathrm{s}(...(\sem{0}))))$ where the second argument
encodes $n$ as $n$-fold ``successor'' of 0. Similar special cases are $\sin(n \times x)$ and $\cos(n \times x)$. These translate into similar terms with $\mathrm{sin\_n}$ and $\mathrm{cos\_n}$, respectively.
These patterns are recognized with type-checking constraints that are evaluated by the host language Python.
Unary minus and division are eliminated in terms with multiplication by $-1$ and 
exponentiation by $-1$, respectively. \NORM also replaces every number $n$
by $\sem{n}$, and every parameter $\para a$ by $\sem{1}\times\para{a}\expsym\sem{1}$.

The \CANON and \SIMP rewrite systems are defined in Tables~\ref{tab:canon-reformatted} and
\ref{tab:simp-reformatted}, respectively. Their A1.2, A1.5, S1 and S2 rules have term ordering constraints of the
form $x \succ y$ between variables. As an outlier, the rule T1.7 is the \emph{only} rule not oriented with our
term ordering. 
This
circumstance did not lead to non-termination in our experiments but should certainly be addressed rigorously.

\begin{table}[h]
    \centering
    \caption{\CANON rewrite rules}
    \label{tab:canon-reformatted}
    \resizebox{\columnwidth}{!}{%
    \begin{tabular}{l l c l c}
        \hline
        A1.1 & $(x + y) + z$ & $\rightarrow$ & $x + (y + z)$ &  \\
    \hline
    A1.2.1 & $x + y$ & $\rightarrow$ & $y + x$ & $x\succ y$  \\
    A1.2.2 & $x + (y + z)$ & $\rightarrow$ & $y + (x + z)$ & $x\succ y$  \\
    \hline
    A1.3.1 & $\sem{0} + x$ & $\rightarrow$ & $x$ &  \\
    A1.3.2 & $\sem{a} + \sem{b}$ & $\rightarrow$ & $\sem{a + b}$ &  \\
    A1.3.3 & $\sem{a} + (\sem{b} + z)$ & $\rightarrow$ & $\sem{a + b} + z$ &  \\
    A1.3.4 & $(\sem{a} \times x) + (\sem{b} \times x)$ & $\rightarrow$ & $\sem{a + b} \times x$ \\ 
    A1.3.5 & $(\sem{a} \times x) + ((\sem{b} \times x) + z)$ & $\rightarrow$ & $(\sem{a + b} \times x) + z$ &  \\
    \hline
    A1.4 & $(x \times y) \times z$ & $\rightarrow$ & $x \times (y \times z)$ &  \\
    \hline
    A1.5.1 & $x \times y$ & $\rightarrow$ & $y \times x$ & $x\succ y$  \\
    A1.5.2 & $x \times (y \times z)$ & $\rightarrow$ & $y \times (x \times z)$ & $x\succ y$ \\
    \hline
    A1.6.1 & $\sem{0} \times x$ & $\rightarrow$ & $\sem{0}$ &  \\
    A1.6.2 & $\sem{a} \times \sem{b}$ & $\rightarrow$ & $\sem{a \times b}$ &  \\
    A1.6.3 & $\sem{a} \times (\sem{b} \times z)$ & $\rightarrow$ & $\sem{a \times b} \times z$ &  \\
    \hline
    A1.7.1 & $x \times (y + z)$ & $\rightarrow$ & $(x \times y) + (x \times z)$ &  \\
    A1.7.2 & $(y + z) \times x$ & $\rightarrow$ & $(y \times x) + (z \times x)$ &  \\
    \hline
    A1.8.1 & $(x \expsym y) \expsym z$ & $\rightarrow$ & $x \expsym (y \times z)$ &  \\
    A1.8.2 & $(x \expsym y) \times (x \expsym z)$ & $\rightarrow$ & $x \expsym (y + z)$ &  \\
    A1.8.3 & $(x \expsym y) \times ((x \expsym z) \times v)$ & $\rightarrow$ & \rlap{$(x \expsym (y + z)) \times v$} &  \\
    \hline
    A1.9.1 & $\sem{a} \expsym \sem{b}$ & $\rightarrow$ & $\sem{a \expsym b}$ &  \\
    A1.9.2 & $\fun{pwr\_n}(x,\sem{0})$ & $\rightarrow$ & $\sem{1}$ &  \\
    A1.9.3 & $\fun{pwr\_n}(x,\fun{s}(n))$ & $\rightarrow$ & $x \times \fun{pwr\_n}(x,n)$ &  \\
    A1.9.4 & $(x + y) \expsym \sem{1}$ & $\rightarrow$ & $x + y$ &  \\
    A1.9.5 & $(x \times y) \expsym \sem{a}$ & $\rightarrow$ & $(x \expsym \sem{a}) \times (y \expsym \sem{a})$ &  \\
    A1.9.6 & $x \expsym \sem{0}$ & $\rightarrow$ & $\sem{1}$ &  \\
    \hline
    T1.1 & $\sin(\sem{-1} \times x)$ & $\rightarrow$ & $\sem{-1} \times \sin(x)$ &  \\
    T1.2 & $\cos(\sem{-1} \times x)$ & $\rightarrow$ & $\cos(x)$ &  \\
    T1.3 & $\sin(x_1 + x_2)$ & $\rightarrow$ & $\sin(x_1) \times \cos(x_2) +$ & \\ 
    & &  & $\cos(x_1) \times \sin(x_2)$ & \\
    \hline
    T1.4 & $\cos(x_1 + x_2)$ & $\rightarrow$ & $\cos(x_1) \times \cos(x_2) $& \\
    & & &$+ \sem{-1}\sin(x_1) \times \sin(x_2)$ &  \\
    \hline
    T1.5 & $\fun{cos\_n}(\fun{s}(\fun{s}(n)),x)$ & $\rightarrow$ & $\sem{2} \times (\cos(x) \times \fun{cos\_n}(\fun{s}(n),x))$ &  \\ 
    & & &$+ (\sem{-1} \times \fun{cos\_n}(n,x))$ &  \\
    \hline
    T1.6 & $\fun{sin\_n}(\fun{s}(\fun{s}(n)),x)$ & $\rightarrow$ & $\sem{2}\cos(x) \times \fun{sin\_n}(\fun{s}(n),x)$ & \\ 
    & & & $+ (\sem{-1} \times \fun{sin\_n}(n,x))$ &  \\
    \hline
    T1.7 & $\sin(x)\expsym \sem{2}$ & $\rightarrow$ & $\sem{1} + \sem{-1}\times \cos(x)\expsym \sem{2}$ & Not oriented\\
    \hline
    T2.1 & $\fun{cos\_n}(\fun{s}(\sem{0}),x)$ & $\rightarrow$ & $\cos(\sem{1} \times x)$ &  \\
    T2.2 & $\fun{cos\_n}(\sem{0},x)$ & $\rightarrow$ & $\cos(\sem{0})$ &  \\
    T2.3 & $\fun{sin\_n}(\fun{s}(\sem{0}),x)$ & $\rightarrow$ & $\sin(\sem{1} \times x)$ &  \\
    T2.4 & $\fun{sin\_n}(\sem{0},x)$ & $\rightarrow$ & $\sin(\sem{0})$ &  \\
    T2.5 & $\sin(\sem{x})$ & $\rightarrow$ & $\sem{\sin(x)}$ &  \\
    T2.6 & $\cos(\sem{x})$ & $\rightarrow$ & $\sem{\cos(x)}$ &  \\
        \hline
    \end{tabular}%
    }
\end{table}

The \CANON system (see Table \ref{tab:canon-reformatted}) has rules for addition, multiplication, exponentiation,
distributivity, and for evaluating \kwote-terms that are combined by arithmetic operators.
It is not hard to see that every quoted term in every derivable term is initially given as
a number (by \NORM) or will be $\simp${}lified to a number as part of rewriting steps.
\CANON has rules for ``sorting'' factors of products in increasing order wrt.\ $\succ$ ,
for example $x \times (y \times z) \rightarrow y \times (x \times z)\mid x\succ y$. Sorting is important for 
collecting like-terms as then only adjacent terms need to be considered.
\CANON has rules for trigonometric identities, and for expanding exponentiation with
integer constants. For example $(\para a + \para b)\expsym 3$ will be fully multiplied out in
the obvious way.

The main task of \SIMP is to sort sums of monomials so that like-monomials can be
collected, e.g., with $(\sem{a} \times x) + (\sem{b} \times x) \rightarrow \sem{a + b} \times x$ as
one of these rules.\footnote{
\SIMP cannot be integrated with \CANON as \SIMP needs a
right-to-left status for multiplication instead of left-to-right so that leading number coefficients are ignored
for sorting. Their combination into one system leads to problems in proving termination.}

\begin{table}[h]
    \centering
    \caption{\SIMP rewrite system rules.}
    \label{tab:simp-reformatted}
    \resizebox{\columnwidth}{!}{%
    \begin{tabular}{l l c l l}
        \hline
        S1 & $(\sem{a} \times x) + (\sem{b} \times y)$ & $\rightarrow$ & $(\sem{b} \times y) + (\sem{a} \times x)$ & $x\succ y$  \\ 
    S2 & $(\sem{a} \times x) + ((\sem{b} \times y) + z)$ & $\rightarrow$ & $(\sem{b} \times y) + ((\sem{a} \times x) + z)$ & $x\succ y$  \\ 
    S3 & $(\sem{a} \times x) + (\sem{b} \times x)$ & $\rightarrow$ & $\sem{a + b} \times x$ &  \\
    S4 & $(\sem{a} \times x) + ((\sem{b} \times x) + z)$ & $\rightarrow$ & $(\sem{a + b} \times x) + z$ &  \\
        \hline
    \end{tabular}}
\end{table}

Finally, \CLEAN consists of the three rules $\sem{x} \rightarrow x$, $1 \times x \rightarrow x$ and $x \expsym
1 \rightarrow x$ for a more simplified presentation of the final result.

\begin{example}[ARI-normal form]
\label{ex:ARI-normal-form}
  Consider the following ARI-normal form computation $\text{norm}(s) \to_{\ARI} t$ which has parameters
$\para b \succ \para a$. It uses standard
mathematical notation for better readibility; multiplication $\cdot$ is \emph{left} associative. In the example we use $\cdot$ instead of $\times$ to save space.
  \begin{align*}
  \text{norm}(s) & = \text{norm}({{2}\cdot{{\para{b}}\cdot{{3}\cdot{{\para{a}}\cdot{{5}\cdot{\para{b}}}}}}}+{5}) \tag{1}\\
  \to_{\NORM}^*\  &  {{\sem{2}}\cdot{{{\sem{1}}\cdot{{\para{b}}^{\sem{1}}}}\cdot{{\sem{3}}\cdot{{{\sem{1}}\cdot{{\para{a}}^{\sem{1}}}}\cdot{{\sem{5}}\cdot{{\sem{1}}\cdot{{\para{b}}^{\sem{1}}}}}}}}}+{\sem{5}}\tag{2}\\
 \to_{\CANON}^*\ & {\sem{5}}+{{\sem{6}}\cdot{{\sem{5}}\cdot{{\sem{1}}\cdot{{{\sem{1}}\cdot{{\para{a}}^{\sem{1}}}}\cdot{{{\sem{1}}\cdot{{\para{b}}^{\sem{1}}}}\cdot{{\para{b}}^{\sem{1}}}}}}}}\tag{3}\\
  \to_{\CANON}^*\ & {\sem{5}}+{{\sem{30}}\cdot{{{\para{a}}^{\sem{1}}}\cdot{{\para{b}}^{\sem{2}}}}}\tag{4}\\
  \to_{\SIMP}^*\ & {\sem{5}}+{{\sem{30}}\cdot{{{\para{a}}^{\sem{1}}}\cdot{{\para{b}}^{\sem{2}}}}} \tag{5}\\
  \to_{\CLEAN}^*\ & {5}+{{30}\cdot{{\para{a}}\cdot{{\para{b}}^{2}}}}\tag{6} = t
  \end{align*}
  Line (2) demonstrates the replacements (a) and (b) mentioned above with \NORM.
  It achieves that every monomial over parameters is either a number or a product of (at least
  one) number and parameters with exponents. This form is assumed and exploited by the
  \CANON rules by moving all numbers to the left with aggregated multiplication, and
  sorting all parameters with exponents in increasing order wrt.\ $\succ$.
  The ordering $\succ$ is  such that every \kwote-term is smaller than every
  non-\kwote term, e.g., $\para a \succ \sem{1}$. 
  Line (3) is a
  snapshot of the \CANON process before reaching \CANON-normal form on line (4). Notice that two
  occurrences of the $\para b\expsym\sem{1}$ term have been like-collected into $\para
  b\expsym\sem{2}$. The \SIMP rewrite system has no effect in this example, see line
  (5). Finally, line (6) simplifies and unquotes numbers.
  ARI-normal form computation hence uses quoting only as an intermediate device for
  triggering built-in evaluation of arithmetic terms.

  Some examples for normalization of trigonometric terms:
  \begin{align*}
  \sin({\para{a}}+{\para{b}}) & \to_{\ARI} 
  {{\cos(\para{a})}\cdot{\sin(\para{b})}}+{{\cos(\para{b})}\cdot{\sin(\para{a})}}\\
  \sin({3}\cdot{\para{c}}) & \to_{\ARI}
  {{-1}\cdot{\sin(\para{c})}}+{{4}\cdot{{\cos(\para{c})}\cdot{{\cos(\para{c})}\cdot{\sin(\para{c})}}}}\\
  \sin({{\para{\pi}}/{2}}-{\para{\phi}}) & \to_{\ARI} {} \\
  & \rlap{${{-1}\cdot{{\sin(\para{\phi})}\cdot{\cos({0.5}\cdot{\para{\pi}})}}}+{{\cos(\para{\phi})}\cdot{\sin({0.5}\cdot{\para{\pi}})}}$}
  \end{align*}
  
  \end{example}

\begin{note}[Uniqueness and confluence]
    The intention behind \CANON is to sort the parameters with exponents by their bases
    only, for collecting like-terms. 
 Unfortunately, this is not always possible. For example, 
 $\para{b}^{\para{d}} \succ {\para a}^{\para e}$
 is sorted as intended, where ${\para b} \succ {\para a}$.
 However, 
 ${\para a}^{\para{b}^{\para{d}}} \succ \para{b}^{\para{d}}$ is not. This phenomenon can
 lead to non-confluence. Notice, it required an exponent $\para{b}^{\para{d}}$
 that is equal to (or greater than) another factor. Luckily, such cases are rare in physics
 exams. 
\end{note}
 
\begin{theorem}[Soundness of ARI]
  Every rule in each of the rewrite systems \NORM, \CANON, \SIMP and \CLEAN is sound.
\end{theorem}

\subsection{Termination of Rewriting}
\label{sec:termination}
A rewrite system $R$ is \define{terminating} if there is no infinite rewrite
derivation. That is, there is no sequence
of one-step rewrites $t \to_R t_1 \to_R t_2 \to_R \cdots$, for any term $t$.
A standard way to prove (standard) rewrite systems terminating over a finite signature is
to define a \emph{reduction ordering $\succ$} 
such that all rules are \define{oriented}, i.e., $l \succ r$ for every standard rule $l \to r$.
The counterpart for our constrained rules is to show that
every rule $\rho$ (in ARI) is \define{oriented wrt.\ ordinary instances (for $\succ$)}, i.e., that every ordinary instance
$s \Rightarrow t$ of $\rho$ satisfies $s \succ t$. This requires specific analysis for each
constrained rule. For example,  $x+y \to y+x\mid x \succ y$ is oriented in this sense if
the symbol ``$+$'' has left-to-right lexicographic status: it holds that, for every ordinary instance $s+t \Rightarrow t+s$, where $s \succ t$, the term $s+t$ is greater than $t+s$ wrt. $\succ$. The ordinary instance is not oriented if the order in the tuple were reversed, i.e.,
if ``$+$'' had right-to-left status.

We are going to define our term ordering $\succ_{\AWPO(X)}$ as both an extension (by
\kwote-terms) and instantiation of a weighted path ordering
(WPO)~\cite{yamada_unified_2015} where $X$ is a precedence on the
variables which is needed for termination analysis and for confluence analysis.

\paragraph{WPO.}
A \define{reduction ordering $\succ$} is a strict partial well-founded ordering on ($\Sigma_\Pi$-)terms that is
\define{stable (under substitution)}, i.e., if $s \succ t$ then $s\delta \succ t\delta$ for every substitution
$\delta$, and \define{monotonic}, i.e., if $s \succ t$ then $u_p[s] \succ u_p[t]$ for every term $u$.

Following \cite{yamada_unified_2015}, a weighted path ordering \define{WPO} has two parameters: a
quasi-precedence $\succsim_{\calF}$ on a finite signature $\calF$ of function symbols,
and a well-founded $\calF$-algebra $\calB$ equipped with strict and non-strict partial
orders $>$ and $\ge$ on its carrier set.
We use the letter $\calB$ instead of Yamada's $\calA$ in order to avoid confusion with our $\calA_\pi$.
WPOs are denoted as $\succ_{\WPO(\calB,\sigma)}$ where $\sigma$ is a status (for permuting tuples for
lexicographic ordering). 

Informally, to check if $s \succ_{\WPO(\calB,\sigma)} t$ holds one first computes a ``weight'' for
$s$ and $t$ using $\calB$. If the weight of $s$ is $>$ than that of $t$ then the answer is
``yes''. Otherwise, if the weight of $s$ is $\ge$ than that of $t$ then the answer is given by resorting
to the lexicographic path ordering (LPO) with base ordering $\succ_{\WPO(\calB,\sigma)}$ itself.
For non-ground $s$ and $t$ the weights can be symbolic expressions
over the variables of $s$ and $t$, and comparing weights then requires theorem proving of
universally quantified ordering constraints. We
use Z3 for this. 
If some additional conditions are met (``weak monotonicity'' and ``weak simplicity'') then the WPO is a reduction ordering, as desired.

Let $X = (x_1 > x_2 > \cdots > x_n)$ be a finite list of pairwise different variables (from $V$) in decreasing
precedence, possibly the empty list $\varepsilon$. 
We specify our signature $\calF$ for the WPO with a parameter for $X$ because $X$ will
vary.
The entries of $X$ stand for unknown ground terms with an externally provided order among them.
Let $\calF(X) = X \cup \Sigma_\Pi \cup \{t\mid \text{$t$ is a parameter-free and quote-free term}\}$ be
the signature for our WPO (here $X$ is a set).
We fix a precedence $\succsim_{\calF(X)}$ that consists of two separate strands, (1) $X$, and
(2) $\sin > \cos > \expsym > \times > + > p_1 > \cdots > p_n > \kwote > t$
for every parameter-free and quote-free term $t$, where $\Pi = \{p_1, \ldots, p_n \}$ and 
the order of the $p_i$s is arbitrary but fixed.
We treat the variables $x_i$ as if they were constants in the LPO part of the WPO so that
a total ordering on $X$ results. (All other variables could similarly be treated as constants without
any precedences.)

The algebra $\calB$ comes with a weight function $w$ from the terms
to ``arithmetic'' parameter-free and \kwote-free terms as follows (math notation for readability):
\begin{align*}
w(v) & = v \text{, for variables $v \in V$ }\\
w(\mathrm{cos\_n}(n, x)) & = w(\mathrm{sin\_n}(n, x)) = {({{2}\cdot{{2}^{{w(x)}\cdot{3}}}}+{1})}^{{2}\cdot{w(n)}}\\
w(\mathrm{pwr\_n}(x, n)) & = {({{2}^{w(n)}}+{1})}^{w(x)}\\
w(\mathrm{s}(n)) &= w(n) + 1\\
  w(x\expsym y) & = {({w(x)}\cdot{w(x)})}\cdot{({2}\cdot{w(y)})} \\
w(x\times y) & = {2}\cdot{{w(x)}\cdot{w(y)}}\\
w(x+ y) & = {1} + {{w(x)}+{w(y)}}\\
w(\sin(x)) & = w(\cos(x)) = {2}^{{w(x)}\cdot{3}}\\
w(p) & = 1 \text{, for parameters $p \in \Pi$}\\
  w(\sem{t}) & = w(t) = 1
\end{align*}

Let $\calB$ be the algebra with the natural numbers as carrier set and such that
$\calB(s) > \calB(t)$ iff for all assignments $\alpha$ from $V$ to the natural numbers, $\calB(\alpha)(w(s)) >
\calB(\alpha)(w(t))$, analogously for $\ge$.

Finally, we define the \define{ARI WPO $\succ_{\AWPO(X)}$} as the WPO $\succ_{\WPO(\calB,\sigma)}$ with 
$\calF(X)$, $\succsim_{\calF(X)}$ and $\calB$ as defined above. The status $\sigma$ is usually ``left-to-right''.
If $\succ_{\WPO(\calB,\sigma)}$ is clear from the context we just write $\succ$ for simplicity.

\begin{example}
Let $x$ and $y$ be foreground variables, $a$, $b$ be background variables and $X = (x > y)$. The terms $x$, $\sem{1}$, $\sem{2}$ $\sem{1+1}$, $\sem{a}$, $\sem{a+b}$ are all
incomparable with each other (wrt.\ $\succ$).
However, it holds $x \succ y$, $x+y \succ y+x$, $\sem{a} + \sem{b} \succ \sem{a + b}$, all by
weighing alone. For the last example, we have
$w(\sem{a} + \sem{b}) = (1 + w(a) + w(b)) = (1 + a + b)$ and $w(\sem{a + b}) = 1$.
It is easy to see that every term has a weight $\ge 1$. This allows us to assume $a, b \ge 1$ and conclude $1 + a + b > 1$.
\end{example}

\begin{note}[AWPO is a reduction ordering]
  \label{note:AWPO-well-founded-stable}
  The signature $\calF$ in the definition of AWPO includes the infinitely many $\Sigma_P(V)$-terms as
  constant symbols that are \emph{pairwise unrelated} in the precedence and of lower
  precedence than every function symbol and parameter in $\Sigma_\Pi$.
  Every \kwote-term gets the lowest possible weight of 1. Together it follows that every
  $\Sigma_\Pi$-term is greater than every \kwote-term. This trivially preserves WPO's stability under
  substitution. Because every chain of decreasing precedences is finite,
  well-foundedness of WPO carries over to AWPO.
\end{note}

\begin{theorem}[Termination of ARI]
  The rewrite systems \CANON without rule T1.7 and \SIMP are terminating.
\end{theorem}
\begin{proof}
  All rules in \CANON except T1.7 and \SIMP are oriented wrt.\ ordinary instances for $\succ_{\AWPO(X)}$,
  with the exception of rule T1.7.
  This  can be proven (almost) automatically by applying the method in
  Section~\ref{sec:termination-analysis-overview}. Some manual inspection and filling gaps was
  necessary. Together with well-foundedness (Note~\ref{note:AWPO-well-founded-stable}) the
    theorem follows with standard results. In particular, every rewrite derivation from
    any $\Sigma_\Pi$-term is finite with underlying ordering $\succ_{\AWPO(\epsilon)}$.
\end{proof}

\section{The Dataset}
\label{sec:Dataset}

We examined student responses to the Australian Physics Olympiad exam of 2023. There were a total of 1526 typed student responses (including blank responses) to each of the 45 questions. The marking team's grade for each student response was also provided. 
Approximately 10\% of the responses were hand-written, scanned and attached as images. The hand-written responses were not considered in this study and therefore we changed students scores for these responses to $0$.

We focus on grading two questions. Question 25 requires flexible marking, as there were multiple forms of the correct answer. Question 26 requires students to use trigonometric functions.

\subsection{Question 25}
\label{sec:Q25}
Question 25  is worded as follows:

\small{
\begin{quote}
    \textit{Consider two solid, spherical masses, one with mass $m_1$, and one with mass $m_2$. Assuming that $m_2$ is initially at rest, and that $m_1$ is incident on $m_2$ with some energy $E_0$, the particles will scatter with final energies $E_1$ and $E_2$ respectively (as shown).
    Write an equation for conservation of energy for this process. Only include the following variables in your answer:}
    \begin{itemize}
        \item \textit{$m_1$ – the mass of the incoming particle}
	\item \textit{$m_2$ – the mass of the originally stationary target}
	\item \textit{$E_0$ – the kinetic energy of $m_1$ before the collision}
	\item \textit{$E_1$ – the kinetic energy of $m_1$ after the collision}
	\item \textit{$E_2$ – the kinetic energy of $m_2$ after the collision}
    \end{itemize}
\end{quote}}

\normalsize
\noindent The question has a simple correct answer: $E_0 = E_1 + E_2$. However many students are taught to substitute expressions for kinetic energy ($KE = \frac{mv^2}{2}$) when writing conservation of energy equations. Therefore many students gave answers such as $m_1 v_0^2 = m_1 v_1^2 + m_2 v_2^2$ which are equivalent to the correct answer and markers would be able to determine the physical/algebraic equivalence of this student response. 

To allow our system to mimic this grader behavior, the substitutions in Table \ref{tab:sub} were applied to the student expressions exhaustively as the final step of pre-processing. After this, both Z3 (an SMT solver) and our TRS determined the equivalence of the student answer and the correct solution and assigned a grade to the student. 

\begin{table}[h]
\centering
\caption{Expressions for energy and momentum of particles before and after an interaction}
\label{tab:sub}
\resizebox{\columnwidth}{!}{
\begin{tabular}{rll}
\toprule
Original  & New  & Description \\
Quantity  & Expression & \\
\midrule
$E_0$  $\rightarrow$ & $\frac{m_1v_0^2}{2}$ & Initial kinetic energy of particle 1 \\
$p_0$  $\rightarrow$ & $m_1v_0$             & Initial momentum of particle 1 \\
$E_1$  $\rightarrow$ & $\frac{m_1v_1^2}{2}$ & Final kinetic energy of particle 1 \\
$p_1$  $\rightarrow$ & $m_1v_1$             & Final momentum of particle 1 \\
$E_2$  $\rightarrow$ & $\frac{m_2v_2^2}{2}$ & Final kinetic energy of particle 2 \\
$p_2$  $\rightarrow$ & $m_2v_2$             & Final momentum of particle 2 \\
\bottomrule
\end{tabular}
}
\end{table}

\paragraph{The Z3 Grading process of Question 25} is as follows. 
We define the \define{trigonometric axiom set} $T$ as a set of equations  corresponding to trigonometric axioms where $x$ is a variable implicitly universally quantified within each formula. See Section \ref{sec:IniZ3res} for an example of set $T$. 

Let  $D_i = \{d_{i1}, d_{i2}, \dots, d_{ip} \}$ be the set of equations contained in the $i$th student's response and $C = \{c_1, c_2, \dots,c_q\}$ be the set of equations required in the marking scheme. We generate a set of additional inequalities, which we call \define{non-zero constraints}, $Z = \{z_1, z_2, \dots, z_l\}$ which prevent expressions from being undefined. For example, expressions in denominators are not allowed to equal 0.

We define two equations $d$ and $c$ as \define{z3-equivalent}, $d \equiv_{Z3} c$, if $T \models \forall (Z \to (d \equiv c))$. 
We then break the equivalence checking task into two unsatisfiability checks for Z3. 

For Question 25, $C$ consists only of one equation, $C = \{E_0 = E_1 + E_2\}$ to which the Table \ref{tab:sub} rules are applied. We refer to this equation as $c_1$. The Z3 assigned mark for the $i$th student is given by:
\[
M_{\text{Z3}i} =
\begin{cases}
    1 & \text{if } \exists d_{ip} \in D_i \text{ such that } d_{ip} \equiv_{Z3} c_1 \\
    0 & \text{otherwise}
\end{cases}
\]

\paragraph{The TRS grading procedure for Question 25} was different. We define \define{solving an equation $e$ for a parameter $x$} as the process of performing valid algebraic operations to the equation such that the LHS consists only of $x$ and the RHS does not contain $x$. In general, not all equations are solvable. 

We made the choice to solve all student equations $d_{ij}$ for the parameter $v_0$ because the pre-processing substitutions in Table \ref{tab:sub} guarantee that a correct answer must contain this parameter. Therefore equations were expressed in the form $v_0 = e_j$ where 
$e_j$ is an expression containing numbers, parameters and well founded function symbols but not $v_0$. 
 
We define a \define{solving function, $f$}, such that:
\[
f(d_{ij}) =
\begin{cases}
e_j & \text{if $d_{ij}$ can be solved by our computer}\\ 
& \text{algebra system for $v_0$} \\
\text{NaN} & \text{otherwise}
\end{cases}
\]
Where \define{NaN} is a special expression with the property $\text{NaN} \approx c = \bot$ any expression $c$. See Section \ref{sec:theorem-proving-by-normalization} for definition of algebraically equal, $\approx$. 

We implemented the solving function $f()$ using the SymPy algebra system \cite{SymPy}. We note that there were no cases where SymPy was unable to solve for $v_0$. The rewrite system mark for the $i$th student was given by: 
\[
M_{\text{TRS}i} =
\begin{cases}
    1 & \text{if } \exists d_{ij} \in D_i | f(d_{ij}) \approx f(c_1) \\
    0 & \text{otherwise}
\end{cases}
\]


Intuitively this means that students were awarded a mark if any of the equations that they wrote were equivalent to the correct answer. We note that this may not be an appropriate way to mark complex questions which require students to demonstrate understanding through correct working, however it is sufficient for this one-mark question. 

\subsection{Question 26}
\label{sec:Q26}
Question 26 was worded as follows:

\begin{quote}
    \textit{Write two equations for conservation of momentum, accounting for the scattering angles $\phi$ and $\theta$.}
\end{quote}

\begin{wrapfigure}[8]{r}{0.23\textwidth}
  \hspace{-2em}\includegraphics[width=0.25\textwidth]{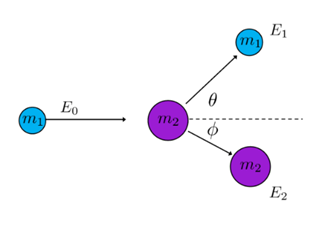}
  \label{fig:Q26Diagram}
\end{wrapfigure} The same LLM pre-processing was applied to student responses to Question 26 to standardize their format and remove syntax errors. 

However as the marking scheme for Question 26 was more complex, a different formula was used to grade the student responses. 

\paragraph{The Z3 Grading Procedure for Question 26} was as follows.
Let $W = \{w_1, w_2, ..., w_q\}$ be the \define{marking weights} corresponding to each required equation $c \in C$. We define the \define{identity function}, $\mathbb{I}(\phi)$ as:
\[
\mathbb{I}(\phi) =
\begin{cases}
    1 & \phi = \top \\
    0 & \phi = \bot
\end{cases}
\]

The Z3 mark assigned to the student is given by:
\[
M_{\text{Z3}i} = \sum_{k=1}^q w_k \cdot \mathbb{I} (\exists d_{ij} \in D_i | d_{ij} \equiv_{Z3} c_k)
\]

Note that for Question 26 $q=2$ and  $w_1=w_2=0.5$. Intuitively, students were awarded 0.5 marks if any of their equations matched the correct x-momentum equation, and 0.5 marks if any of their equations matched the correct y-momentum equation. In Sections \ref{sec:IniZ3res} and \ref{sec:Z3noAxioms} we describe how this technique enabled Z3 to assign the correct marks to most of the student responses. 

\paragraph{The TRS grading procedure for Question 26} was similar. The rewrite system awarded points to the students in Question 26 as follows:
\[M_{\text{TRS}i} = \sum_{k=1}^q w_k \cdot \mathbb{I}(\exists d_{ij} \in D_i | f(d_{ij}) \approx f(c_k))
\]
This means that students were awarded marks according to the weight on the marking scheme for each equation they wrote which was algebraically equivalent to a corresponding article on the scheme. This mirrors the behavior of the majority of markers for most physics questions. Both the term rewriting system and Z3 were applied to these questions and the results are shown in Section \ref{sec:results}.

\section{Results and Discussion}
\label{sec:results}

A series of experiments were performed to evaluate the performance of our TRS using Z3 as a control. The results of these experiments are described in detail in Sections \ref{sec:IniZ3res}, \ref{sec:Z3noAxioms} and \ref{sec:TRSres}. A summary of all of the experimental results can be found in Table \ref{tab:results}. If the marking method failed to assign a grade or assigned a student grade different to the ground truth grade, this was counted as a fail. 

\begin{table}[htbp]
  \centering
  \footnotesize           
  \caption{Summary of experimental results for Questions 25 and 26.}
  \label{tab:results}
  \begin{threeparttable}
    \begin{tabular}{@{}lrrrr@{}}
      \toprule
      \multirow{2}{*}{Method} & \multicolumn{2}{c}{Q25} & \multicolumn{2}{c}{Q26} \\
      \cmidrule(lr){2-3}\cmidrule(lr){4-5}
                              & CPU Time & Number & CPU Time & Number \\
                              & (seconds) & of fails & (seconds) & of fails \\
      \midrule
      Z3 + trig.\           &  550 &  18 & 5000 & 355 \\
      axioms\tnote{a} & & & & \\
      Z3 + custom&   \textbf{16} &   \textbf{0} &   \textbf{40} &   5 \\
      sqrt tactic\tnote{b} & & & &\\
      TRS\tnote{c}                  & 1140 &   \textbf{0} &  612 &   1 \\
      TRS + $T'$ \tnote{d}                  & 1210 &   \textbf{0} &  870 &   \textbf{0} \\
      \bottomrule
    \end{tabular}
    \begin{tablenotes}
      \footnotesize
      \item[a] Fails if student equation contains trigonometric functions or square root symbols. In most of the observed failure cases returns timeout error.
      \item[b] Fails on non-linear combinations of trigonometric expressions and cases requiring angle-addition formulas. In four of the five observed failure cases returns \texttt{unknown} so the user is aware of the fail.
      \item[c] There was a failure case because the system was missing trigonometric identities, resulting in a false negative. 
      \item[d] There were no observed failures after the extra $T'$ axioms were added.
    \end{tablenotes}
  \end{threeparttable}
\end{table}

All experiments were run on a local machine with an Intel Core-i9-13900K CPU (3-5.8GHz), 64GB of DDR5 (4800MT/s) and an NVIDIA GeForce RRTX 4060Ti GPU with 16GB of dedicated GPU memory. 

\subsection{Initial Z3 experiments}
\label{sec:IniZ3res}
One of the first problems when applying Z3 is to find an appropriate trigonometric axiom set $T$ that allows Z3 to understand the trigonometric functions but does not quickly lead to time-out or memory limits. By default we use trigonometric axioms that correspond to our rewrite rules:
\[
\small
T= \begin{cases}
    \sin(-x) &= -\sin(x) \\
    \cos(-x) &= \cos(x) \\
    \sin^2(x) &= 1 - \cos^2(x) \\
    \sin(x_1 + x_2) &= \sin(x_1)\cos(x_2) + \cos(x_1)\sin(x_2) \\
    \cos(x_1 + x_2) &= \cos(x_1)\cos(x_2) - \sin(x_1)\sin(x_2) \\
    \cos((n + 2)x) &= 2\cos(x)\cos((n + 1)x) - \cos(nx)  \\
    \sin((n + 2)x) &= 2\cos(x)\sin((n + 1)x) - \sin(nx) 
\end{cases}
\]

These rules were used to mark the student responses to Question 25. Z3 was able to process the 1526 examples in approximately 10 minutes, successfully classifying all but 18 results. In these cases Z3 was not able to return sat or unsat within a 100 second timeout. Further investigation shows that there were two distinct failure cases.

The first type of failure case occurred when students included trigonometric functions in their responses, for example $m_1v_0=m_1v_1\cos(\theta) + m_2v_2\cos(\phi)$. In these 14 cases the reason Z3 gives for its unknown result is `timeout'. The solver statistics for each of these problems show that the universal quantifiers in the trigonometric rules caused the number of quantified instantiations, added equations and clauses to approach the millions. In these cases the Z3 memory usage quickly grew to tens of gigabytes before hitting hard resource limits.

The second type of failure case occurred when square root symbols or fractional powers were included in student responses. For example:
\[v_2=(((m_1v_0^2) - (m_1v_1^2))/m_2)^{1/2}\]
In these cases, Z3's reason for the unknown result was that the SMT tactic was incomplete, and Z3 failed quickly (less than 0.01 seconds) using less than 20MB of memory. 
Looking at the solver statistics from these cases shows that Z3 attempts to use  Gr\"obner Bases and other Non-Linear Arithmetic (NLA) tools. These tools fail to reduce the problem to a state where Z3 is able to prove unsatisfiability or find a satisfying example. 

The results were similar for Question 26; Z3 gave a timeout error for any expression which contained a trigonometric expression. This resulted in 355 fails. For Z3 to progress it is clear that a simpler set of trigonometric axioms would need to be provided. We note that in each of these examples Z3 fails gracefully, providing an unknown result to alert the user. 

\subsection{Further Z3 Experiments}
\label{sec:Z3noAxioms}
The universally quantified $T$ axioms for the uninterpreted functions $\sin$ and $\cos$ force Z3 to solve a very difficult problem. We performed further experiments where we removed some of the axioms from $T$ to try to improve its grading performance.  

Removing these axioms meant that Z3 would not be able to show the equivalence of some expressions such as $\sin(2\theta)$ and $2\sin(\theta)\cos(\theta)$. However, if most student answers do not require operations to be applied to the arguments of the trig functions then Z3 would be able to correctly grade the responses. 

First we consider applying a reduced set of axioms to Question 25. When the final four of the original seven $T$ axioms are removed, timeout errors no longer occur. However the $\sin^2(x) = 1-\cos^2(x)$ axiom causes the system to fail quickly when student responses contain trig functions. In these cases Z3 recognizes 
that these cases contain non-linear arithmetic and therefore its solvers are incomplete. Removing the third axiom fixes this problem and Z3 was quickly able to provide a sat response for the Question 25 examples that contained trigonometry.  
Note that such (sat) counter examples become increasingly unreliable as axioms are removed. 

\paragraph{Radical Elimination:} After removing the last five axioms, only the four examples which contained square roots caused problems. In these cases, the NLA solver quickly decides that the system is incomplete and gives up. To prevent this, we implement a radical elimination procedure as a pre-processing step. Each square root is replaced with a new variable and two additional constraints, thus converting the problem to be polynomial which can be handed to the NLA solver. For example, consider this student response:
\[E_0=m_1\cos(\theta) + (m_2 - m_1)^{1/2}\]
In this case $(m_2 - m_1)^{1/2}$ is identified as a radical and the system creates a new variable to represent it, $r_0$. Then we remove the original equation from the solver and replace it with the following three: $ \quad
r_0 \geq 0, \quad
r_0 \times r_0 = (m_2 - m_1), \quad
E_0=m_1\cos(\theta) + r_0$.

Once implemented, this allowed Z3 to correctly grade all responses in Question 25. This same approach was applied to Question 26 and resulted in only five fails. The first was that no points were awarded for: $m_1v_0=\sin(\pi/2-\phi)m_2v_2 + \cos(\theta)m_1v_1$. This fail occurred because the angle addition axioms were removed. Note that the following response would have resulted in a similar fail: $m_1v_1= m_1v_1\cos(\theta) + m_2v_2\sin(\pi/2 - \phi)$, except this response was incorrect as it contains a $v_1$ in the place of $v_0$. 

The other four fails were for expressions such as 
\[v_2=(m_2(v_0\cos(\theta))^2 + (v_0\sin(\theta))^2)^{1/2}\] 

This example contains non-linear combinations of uninterpreted functions, a type of problem where Z3 is incomplete. In these cases the solver statistics show that only 20MB of memory  was used and Z3 makes approximately 5 quantifier instantiations with a number of calls to NLA components.  This indicates that in these instances Z3 is not `blowing-up' and exhausting resources, but after creating a few quantifier instantiations stopped because none of them helped progress the proof. 
This is possibly because each new instantiation adds a term that contains an uninterpreted function (trig function) which are treated by the solver as fresh variables. 

Since very few properties of the trigonometric functions were given to Z3, it is unable to solve non-linear combinations of these functions. Overall the reduced axiom set allowed Z3 to quickly find unsat, but as soon as non-identical comparisons of trig functions are required, it lacks the axioms to resolve these instances.  

\subsection{Our TRS Results}
\label{sec:TRSres}
Using the method described in Section \ref{sec:Q25}, our rewrite system was able to correctly grade all 1526 examples for Question 25 in a total CPU time of approximately 1140 seconds. It took 612 seconds for our system to grade Question 26, with one fail occurring. The fail response contained this equation:
\begin{gather*}
    m_1v_0=\sin(\pi/2-\phi)m_2v_2 + \cos(\theta)m_1v_1
\end{gather*}

This equation matches the x-momentum equation. Our rewrite system contains the rule:
\begin{gather*}
    \sin(X_1+X_2) \rightarrow \sin(X_1)\cos(X_2) + \cos(X_1)\sin(X_2).
\end{gather*}
Which should reduce $\sin(\pi/2 - \phi)$ to $\sin(\pi/2)\cos(\phi)-\cos(\pi/2)\sin(\phi)$. However to make the final step $\sin(\pi/2)\cos(\phi)-\cos(\pi/2)\sin(\phi) \rightarrow \cos(\phi)$ would require our system to know that $\sin(\pi/2)=1$ and $\cos(\pi/2)=0$. To solve this issue, the following axioms, $T'$ were added:
\[
T'= \begin{cases}
    \sin(\pi) \rightarrow 0, \hspace{0.2cm}    \sin(\pi/2) \rightarrow 1\\
    \cos(\pi) \rightarrow 1, \hspace{0.2cm}    \cos(\pi/2) \rightarrow 0\\
\end{cases}
\]
Adding these extra trigonometric axioms increased the approximate CPU time to $870$ seconds but allowed our TRS to correctly grade all student responses.

One advantage of our TRS compared to SMT solvers is that adding additional unused axioms only has a minimal impact on the run time. We reran the Question 25 set with the additional four $T'$ axioms, this increased the CPU run-time from  $1140$ seconds to $1210$, an approximately $6\%$ increase. This is a stark contrast to Z3 where adding the additional axioms caused the system to timeout and fail.

\section{Conclusions and Future Work}
\label{sec:limitations}
In this paper, we presented our approach for supporting marking physics exams by combining
LLMs, computer algebra systems, and term rewriting systems.  For the
latter, we designed a non-trivial term rewrite system for the required arithmetic
operations over an expressive rule language with ordering constraints and over infinite
domains. We implemented a normalization procedure for this language, and we developed and
implemented methods for termination and for confluence analysis. This led to theoretical
challenges, whose solutions we sketched in the paper.

Our TRS is able to successfully normalize student equations which contain addition, multiplication, exponentiation and
trigonometric functions.  The TRS approach scales well when redundant rules are added,
with minor penalty in CPU time for normalization. In contrast, the SMT 
solver performance degraded when we added redundant axioms. 
This is not surprising as theorem provers are not known to
always scale well.

\newsavebox{\diverge}
\begin{lrbox}{\diverge}
\begin{minipage}[b]{6.5cm}
  $\begin{tikzcd}[row sep=small, column sep=small]
 & \frac{a+b}{a+b} \arrow[dl, " "'] \arrow[dr, " "] & \\
\frac{a}{a+b} + \frac{b}{a+b} & & 1
\end{tikzcd}
$
\end{minipage}
\end{lrbox}

\noindent \textbf{Some limitations and ideas for future work.} 
Currently the LLM stage of the pipeline is only able to achieve 73\% accuracy, even using techniques which provide the model with feedback. To improve, larger LLMs could be used or models could be fine-tuned to improve performance. 

Our TRS was sufficiently terminating, confluent and complete for grading two 2023 Australian Physics Olympiad problems. 
\begin{wrapfigure}[5]{r}{0.38\columnwidth}
\hspace*{-2em}\raisebox{5ex}[8ex][0ex]{\usebox{\diverge}}
\end{wrapfigure}
However, this is not enough for all physics problems. 
A natural example for expected convergence where our system fails to join is
on the right.

One significant drawback of the TRS is that in general there is no well defined canonical form for arbitrary equations. This means that the TRS is not able to prove that a student's answer is incorrect and leaves the possibility of false negatives. 
A clear example of this is the response that was incorrectly graded in Question 26, requiring the $T'$ axioms to be added. 
In future we will specify the exact functional forms of equations which our TRS can reduce to a unique canonical form, this will define the situations where the system will fail. 

Our TRS could be further improved by adding axioms for complicated expressions involving exponentiation by variables, logarithms and inverse trigonometric functions. 
The scope of problems which can be graded could also be widened, beyond simply checking for algebraic equivalence, to verify student proofs.

Currently our system requires SymPy to first solve the equation for a specific parameter
before applying the TRS.  This means our system is only able to determine the equivalence
of two equations if Sympy is able to solve the equation for at least one variable. This means
that the rewrite system can only be as good as the chosen algebra system's solving capabilities.

Future work will include addressing these limitations. In particular, we want to use our
confluence analysis tool to examine critical pairs and add rules to improve
confluence. Fully automated completion seems problematic. 
We also plan to make the AlphaPhysics TRS more expressive to enable grading of a larger variety of physics problems. 
Finally, our system is implemented in Python, a slow interpreted language, and in a
non-optimized way. Performance was sufficient for our purpose. Re-implementation in a
faster compiled language and/or using efficient data structures,  for example term indexing,
might be an option.

\subsubsection{Acknowledgments}

The authors would like to thank Australian Science Innovations for access to data from the 2023 Australian Physics Olympiad. 
This research was supported by a scholarship from CSIRO's Data61.
The ethical aspects of this research have been approved by the ANU Human Research Ethics Committee (Protocol 2023/1362).


\begin{thebibliography}{}

\bibitem[\protect\citeauthoryear{Abdin and Zhang}{2024}]{abdinPhi4TechnicalReport2024}
Abdin, M. I., and Zhang, Y.
\newblock 2024.
\newblock Phi-4 {{Technical}} {{Report}}.
\newblock Microsoft Research.
\newblock \url{https://www.microsoft.com/en-us/research/publication/phi-4-technical-report/}

\bibitem[\protect\citeauthoryear{Avenhaus and
  Becker}{1994}]{goos_operational_1994}
Avenhaus, J., and Becker, K.
\newblock 1994.
\newblock Operational specifications with built-ins.
\newblock In {\em {{STACS}} 94}. 
\newblock  263--274.

\bibitem[\protect\citeauthoryear{Baader and Nipkow}{1998}]{Baader1998Term}
Baader, F., and Nipkow, T.
\newblock 1998.
\newblock {\em Term {{Rewriting}} and {{All That}}}.
\newblock Cambridge University Press.

\bibitem[\protect\citeauthoryear{Baumgartner and
  Waldmann}{2013}]{baumgartner_hierarchic_2013}
Baumgartner, P., and Waldmann, U.
\newblock 2013.
\newblock Hierarchic {{Superposition}} with {{Weak Abstraction}}.
\newblock In Bonacina, M.~P., ed., {{CADE-24}},  39--57.
\newblock Springer.

\bibitem[\protect\citeauthoryear{Bj{\o}rner and
  Nachmanson}{2024}]{gurfinkel_arithmetic_2024}
Bj{\o}rner, N., and Nachmanson, L.
\newblock 2024.
\newblock Arithmetic {{Solving}} in {{Z3}}.
\newblock In Gurfinkel, A., and Ganesh, V., eds., {\em {{CAV}}}, volume 14681. Springer.
\newblock  26--41.

\bibitem[\protect\citeauthoryear{Chen and Wan}{2025}]{Chen2025Grading}
Chen, Z., and Wan, T.
\newblock 2025.
\newblock Grading explanations of problem-solving process and generating
  feedback using large language models at human-level accuracy.
\newblock {\em Physical Review Physics Education Research} 21(1):010126.

\bibitem[\protect\citeauthoryear{Department of Industry Science and Resources}{2024}]{AustraliaAIEthics2024}
Department of Industry Science and Resources.
\newblock 2024.
\newblock Australia’s {{AI}} {{Ethics}} {{Principles}}.

\bibitem[\protect\citeauthoryear{Dershowitz and
  Jouannaud}{1990}]{dershowitz_chapter_1990}
Dershowitz, N., and Jouannaud, J.-P.
\newblock 1990.
\newblock {{Rewrite Systems}}.
\newblock In {\em Formal {{Models}} and
  {{Semantics}}}, Handbook of {{Theoretical Computer Science}}. 
  Elsevier.
\newblock  243--320.


\bibitem[\protect\citeauthoryear{Harrison}{2009}]{Harrison2009Handbook}
Harrison, J.
\newblock 2009.
\newblock {\em Handbook of {{Practical Logic}} and {{Automated Reasoning}}}.
\newblock Cambridge University Press.

\bibitem[\protect\citeauthoryear{Kambhampati \bgroup et al\mbox.\egroup
  }{2024}]{Subbarao2024LLMmodulo}
Kambhampati, S.; Valmeekam, K.; Guan, L.; Verma, M.; Stechly, K.; Bhambri, S.;
  Saldyt, L.~P.; and Murthy, A.~B.
\newblock 2024.
\newblock Position: {{LLMs Can}}'t {{Plan}}, {{But Can Help Planning}} in
  {{LLM-Modulo Frameworks}}.
\newblock In {\em {{ICML}}},  22895--22907.
\newblock PMLR.

\bibitem[\protect\citeauthoryear{Kaplan and Choppy}{1989}]{goos_abstract_1989}
Kaplan, S., and Choppy, C.
\newblock 1989.
\newblock Abstract rewriting with concrete operators.
\newblock In {\em Rewriting {{Techniques}} and
  {{Applications}}}, volume 355. Springer.
\newblock  178--186.

\bibitem[\protect\citeauthoryear{Kop and Nishida}{2013}]{kop_term_2013}
Kop, C., and Nishida, N.
\newblock 2013.
\newblock Term {{Rewriting}} with {{Logical Constraints}}.
\newblock In {\em
  {{FroCoS}}},  343--358.
\newblock Springer.

\bibitem[\protect\citeauthoryear{Kortemeyer, N{\"o}hl, and
  Onishchuk}{2024}]{Kortemeyer2024Grading}
Kortemeyer, G.; N{\"o}hl, J.; and Onishchuk, D.
\newblock 2024.
\newblock Grading assistance for a handwritten thermodynamics exam using
  artificial intelligence: {{An}} exploratory study.
\newblock {\em Physical Review Physics Education Research} 20(2):020144.

\bibitem[\protect\citeauthoryear{Kortemeyer}{2023}]{Kortemeyer2023Toward}
Kortemeyer, G.
\newblock 2023.
\newblock Toward {{AI}} grading of student problem solutions in introductory
  physics: {{A}} feasibility study.
\newblock {\em Physical Review Physics Education Research} 19(2):020163.

\bibitem[\protect\citeauthoryear{Martin and Nipkow}{1990}]{Martin1990Ordered}
Martin, U., and Nipkow, T.
\newblock 1990.
\newblock Ordered rewriting and confluence.
\newblock In Stickel, M.~E., ed., {\em CADE-10},  366--380.
\newblock Berlin, Heidelberg: Springer.

\bibitem[\protect\citeauthoryear{McGinness and
  Baumgartner}{2025a}]{mcginness_can_2025}
McGinness, L., and Baumgartner, P.
\newblock 2025a.
\newblock Can {{Large Language Models Correctly Interpret Equations}} with
{{Errors}}?
\newblock arXiv:2505.10966 [physics.ed-ph]

\bibitem[\protect\citeauthoryear{Meurer \bgroup et al\mbox.\egroup
  }{2024}]{SymPy}
Meurer, A.; Smith, C.; Paprocki, M.; and Scopatz, A.
\newblock 2024.
\newblock {{SymPy}}: Symbolic computing in {{Python}}.
\newblock \url{https://www.sympy.org/en/index.html}.

\bibitem[\protect\citeauthoryear{Mok \bgroup et al\mbox.\egroup
  }{2024}]{mok_using_2024}
Mok, R.; Akhtar, F.; Clare, L.; Li, C.; Ida, J.; Ross, L.; and Campanelli, M.
\newblock 2024.
\newblock Using {{AI Large Language Models}} for {{Grading}} in {{Education}}:
  {{A Hands-On Test}} for {{Physics}}.
\newblock arXiv:2411.13685 [physics.ed-ph]
  
\bibitem[\protect\citeauthoryear{Ogg}{2024}]{Ogg2024Brisbane}
Ogg, M.
\newblock 2024.
\newblock Brisbane {{AI}} edtech {{Edexia}} accepted into {{Y Combinator}}.
\newblock \url{http://www.businessnewsaustralia.com.html}.

\bibitem[\protect\citeauthoryear{Ramesh and
  Sanampudi}{2022}]{Ramesh2021Automated}
Ramesh, D., and Sanampudi, S.~K.
\newblock 2022.
\newblock An automated essay scoring systems: A systematic literature review.
\newblock {\em Artificial Intelligence Review} 55(3):2495--2527.

\bibitem[\protect\citeauthoryear{Shostak}{1984}]{shostakDecidingCombinationsTheories1984}
Shostak, R.~E.
\newblock 1984.
\newblock Deciding {{Combinations}} of {{Theories}}.
\newblock {\em J. ACM} 31(1):1--12.

\bibitem[\protect\citeauthoryear{Weegar and
  {Idestam-Almquist}}{2024}]{Weegar2024Reducing}
Weegar, R., and {Idestam-Almquist}, P.
\newblock 2024.
\newblock Reducing {{Workload}} in {{Short Answer Grading Using Machine
  Learning}}.
\newblock {\em International Journal of Artificial Intelligence in Education}
  34(2):247--273.

\bibitem[\protect\citeauthoryear{Windle \bgroup et al\mbox.\egroup
  }{2022}]{Windle2022Teachers}
Windle, J.; Morrison, A.; Sellar, S.; Squires, R.; Kennedy, J.; and Murray, C.
\newblock 2022.
\newblock {\em Teachers at Breaking Point}.
\newblock University of South Australia.

\bibitem[\protect\citeauthoryear{Yamada, Kusakari, and
  Sakabe}{2015}]{yamada_unified_2015}
Yamada, A.; Kusakari, K.; and Sakabe, T.
\newblock 2015.
\newblock A unified ordering for termination proving.
\newblock {\em Science of Computer Programming} 111:110--134.

\end{thebibliography}

\appendix

\clearpage\newpage

\appendix
\section{Appendix}

\subsection{Termination Analysis}
\label{sec:termination-analysis-overview}


\begin{lemma}
\label{lemma:stability}
Assume $l \succ_{\AWPO(x>y)} r$ .
If $x\delta \succ_{\AWPO(\epsilon)} y\delta$ then $l\delta \succ_{\AWPO(\epsilon)} r\delta$.
\end{lemma}
\begin{proof}
  (Sketch.)
First we analyse the cases why $l \succ_{\AWPO(x>y)} r$ can hold in terms of the definition
of WPO. In the first case, the ``weighting is $>$'' case, and the ``weighting is $\ge$ and
LPO component holds'' case. For the LPO component case we get an and-or decision tree whose
leaves involving $x$ or $y$ can be precedence facts $x > y$ or encompassment cases between $t[x]$ and
$x$,  which is simpler. The assumption that $x$ and $y$ are not in
precedence relation with other symbols is crucial for that conclusion.

To prove the lemma statement assume $x\delta \succ_{\AWPO(\epsilon)} y\delta$.
Now, $x\delta \succ_{\AWPO(\epsilon)} y\delta$ entails $w(x\delta) \ge w(y\delta)$, $w(x\delta) \ge 1$ and  $w(y\delta) \ge 1$
by the same arguments as said above in the termination procedure.
In  the ``weighting is $>$''  the first case analysis provides us with the validity of the formula
$\forall x, y: (x \ge y \wedge x \ge 1 \wedge y\ge 1 \Rightarrow w(l) > w(r))$
by arguments made above in the termination procedure.
Then by substituting $x\delta$ for $x$, and $y\delta$ for $y$ and modus ponens conclude $w(l\delta) >
w(r\delta)$ as desired.

Otherwise we only get $w(l\delta) \geq w(r\delta)$ by the ``weighting is $\ge$ and LPO component holds''.
In this case, the LPO proof for $l \succ_{\AWPO(x>y)} r$ is replayed with
$l\delta$ and $r\delta$ instead. This results in leafs $x\delta > y\delta$ replacing $x > y$.
Now replace the leaf $x\delta > y\delta$ by the assumed LPO proof of $x\delta \succ_{\AWPO(\epsilon)} y\delta$. This also
discharges the dependency on $x>y$, giving a proof of $l\delta \succ_{\AWPO(\epsilon)} r\delta$ as desired.
\end{proof}

For proving termination of a rewrite system it suffices to show that each of its rules is
oriented wrt.\ ordinary instances.
We provide a high-level overview of our procedure for testing if a rule is 
oriented wrt.\ ordinary instances. It only applies to \CANON
and \SIMP, as these have ordering constraints. The \NORM and \CLEAN systems are
unproblematic.

Let $\rho = (l\to r \mid C)$ be a rule in \CANON or \SIMP.
If $C = \emptyset$ then the procedure tests
$l \succ_{\AWPO(\epsilon)} r$. 
If yes, then $\rho$ is oriented wrt.\ ordinary instances.
This follows from stability of WPO under substitution.
If $C \neq \emptyset$ then $C$ is the singleton $x \succ y$ as $\rho$ is one of the rules from A1.2, A1.5, S1 or S2.

If $C \neq \emptyset$ then $\rho$ is the singleton $x \succ y$. We consider only this shape below. In the general
case of Boolean combinations of ordering constraints, $C$ can have any number of variables. The
procedure has to try all their permutations and also identifications of some of them. For
every such case, the procedure has to test if $C$ is 
\emph{consistent} with the case turned into a precedence. If so, the case is relevant.
This is the same method as explained in the confluence test in the supplementary material in some more detail.

The procedure tests if $l \succ_{\AWPO(x>y)} r$ holds. If yes, then $\rho$ is oriented wrt.\
ordinary instances.
This is because if $x\delta \succ_{\AWPO(\epsilon)} y\delta$ holds for any $\delta$, then $l\delta \Rightarrow r\delta$ is an ordinary instance of $\rho$.
By Lemma~\ref{lemma:stability} $l\delta \Rightarrow r\delta$ ordered.

In the following we explain how the test for $l \succ_{\AWPO(x>y)} r$  is actually carried
out. (Again, it can be generalized.) 
Notice that the rule condition $x\succ y$ becomes the precedence $x>y$, but additionally $x\succ
y$ needs to inform weight computation for effectiveness in practice.

The test follows the structure of the WPO definition and 
first seeks to confirm $l \succ_{\AWPO(x>y)} r$ by proving $w(l) > w(r)$. 
To see how, observe that every variable $x$ in $w(l)$ or $w(r)$ was originally a variable in $\rho$ and 
represents some unknown formula $s$. The minimum weight of every formula is 1, by
construction. The proof of $w(l) > w(r)$ can hence assume $x \ge 1$.
Similarly, let the variables $x$ and $y$ stand for unknown terms $s$ and $t$, respectively,
such that $s \succ_{\AWPO(\epsilon)} t$. The proof of $w(l) > w(r)$ can hence assume $x \ge y$.

The proof obligation can therefore be written as a first-order logic formula: $\forall x, y: (x
\ge y \wedge x \ge 1 \wedge y\ge 1 \Rightarrow w(l) > w(r))$.  The formula is sent to an automated theorem prover
for a proof of validity in the theory of the integers with standard interpretations. We
use Z3, which has the required arithmetic operators readily built-in, but is incomplete in this domain. 
If Z3 succeeds in finding a proof, then $l \succ_{\AWPO(x>y)} r$ holds.
Otherwise $\forall x, y: (x \ge y \wedge x \ge 1 \wedge y\ge 1 \Rightarrow w(l) \ge w(r))$ is tried in a similar way. 
If Z3 succeeds in finding a proof, the LPO component of $\succ_{\AWPO(x>y)}$ is evaluated to
check if $l \succ_{\AWPO(x>y)} r$ holds.
\qed

Using the termination procedure above we could automatically establish orientedness wrt.\
ordinary instances for most rules. In the remaining cases Z3 did not report a conclusive
result for weighting. We completed the analysis affirmatively by proving, by hand, the
validity of the strict weighing constraints.

In the following we show how to make $\succ_{\AWPO}(X)$ ground-total while preserving termination of ARI.

\subsection{Making AWPO ground total}
  \label{sec:ground-totality}
In contrast to WPO, AWPO is not total on ground terms as all (enumerably many)
\kwote-terms are pairwise unrelated. As a result, ground rules can be un-oriented. 
For example, the rule $\rho = \para f(\sem{x}) \to f(\sem{x-1})$
is not oriented with AWPO. Hence its two argument terms are not provably equal, resulting in 
ground incompleteness.

Any AWPO $\succ$ can be made ground total by adding any fixed precedence between numbers, say, using
their natural ordering. Let $\succ^t$ denote that ground-total extension.
With $\succ^t$ we get $0 \succsim_{\calF} -1
\succsim_{\calF} -2 \succsim_{\calF} \cdots$. However, $\{\rho\}$ is not terminating with $\succ^t$, as the infinite
rewriting derivation $f(0) \to_\rho f(-1) \to_\rho f(-2) \to_\rho \cdots$ exists. 
Fortunately, this dilemma is resolved when all rules in a rewrite system $R$ are
\define{well-behaved} in the following sense: every rule $\rho = l \to r\mid C$ in $R$ satisfies the following
property:
\begin{enumerate}
\item[(a)]  $\rho$ is oriented wrt.\ ordinary instances for $\succ$, \emph{or}
\item[(b)] $\rho$ is oriented wrt.\ ordinary  instances for $\succ^t$ and \\
  for every ground ordinary instance $l\gamma \Rightarrow r\gamma$ of $\rho$,
  \begin{enumerate}
  \item[(b1)]  every \kwote-term in $r\gamma$ also occurs in $l\gamma$, and
    \item[(b2)] for every ground term $s$ and $t = t_p[l\gamma]$, if $s \succ t_p[l\gamma]$ then $s \succ t_p[r\gamma]$.
  \end{enumerate}
\end{enumerate}

Condition (b1) makes sure that ground rewrite steps with $\rho$ do not introduce new \kwote-terms.
Condition (b2) makes sure that sequences of rewrite steps of type (b) keep being dominated
dominated by the same term.

Well-behavedness of $\rho$ can be checked sufficiently as a small extension of our
termination analysis procedure in Section~\ref{sec:termination-analysis-overview}:
if $C = \emptyset$ then (a) holds (or not) without
any change. If $C \neq \emptyset$ then the procedure only gives us that $\rho$ is oriented wrt.\
ordinary ground instances for $\succ^t$ (or not). This can be verified by inspecting the
procedure's correctness proof. 
More precisely, the proof of Lemma~\ref{lemma:stability} has one dependence on the symbol
precedences which has to refer to the precedences of $\succ^t$ now.

If $\rho$ is oriented wrt.\ ordinary instances, (b1) and (b2) have to be checked.
A sufficient criterion for (b1) is that every \kwote-term in $r$ also occurs in $l$, which
is easy to automate. 
A sufficient criterion for (b2) is that the set of positions of \kwote-terms in $l\gamma$ and
$r\gamma$ are the same. This entails (b2) because all \kwote-terms have identical properties
for the ordering $\succ$. 

All rules in \CANON and \SIMP are well-behaved according to this check.
For example, $\sem{x}+\sem{y} \to \sem{x+y}$
satisfies (a), and $x+y \to y+x\mid x \succ y$ satisfies (b). Indeed, the latter rule has 
ordinary ground instances, e.g.,  $\sem{1}+\sem{0} \Rightarrow \sem{0}+\sem{1}$ for $\succ^t$, as
$\sem{1}\succ^t\sem{0}$, that do not exist for $\succ$ (as $\sem{1} \not \succ\sem{0}$).
By contrast, the rule $\para f(\sem{x}) \to f(\sem{x-1})$ from above is not well-behaved, it
fails (b1).

Let $R$ be a rewrite system of well-behaved rules. Then $R$ is terminating. This can be seen as
follows. Suppose, by contradiction, an infinite rewrite
derivation $\omega = s_0 s_1 \cdots s_n \cdots$ with rules from $R$.

First, $\omega$ cannot consist of type
(a) steps only because $\succ$ is well-founded.

Second, every subsequence $\omega^{\mathrm b}$ of $\omega$  of contiguous type (b) steps
is finite. This follows from (b1): none of the steps in $\omega^{\mathrm b}$ introduces a new number. This has
the same effect as restricting $\succ^t$ to the finitely many number symbols occurring in
$\omega^{\mathrm b}$. By well-foundedness of WPO, $\omega^{\mathrm b}$ is finite.

Together this means that $\omega$ consists of alternating type (a) and type (b) subsequences.
Consider any maximal type (a) subsequence. It has at least one step, hence at least two
terms. Its last two terms are, say, $s_{n-1}, s_n$ and
it holds  $s_{n-1} \succ  s_n$. Now, follow the subsequent maximal type (b)
sequence, say, $s_n s_{n+1} \cdots s_m$.  By applying property property (b2) repeatedly we get $s_n \succ
s_m$. In words, $s_n \succ s_m$ acts as a bridge to connect the first type (a) sequences with
the next type (a) sequence after the connecting type (b) sequence. We can do this
construction along the whole rewrite derivation. This results in an infinite chain of
decreasing terms wrt.\ $\succ$. This is a contradiction to the well-foundedness of $\succ$.



With the above proof in place, we can now work with ground-total
extensions of AWPOs as orderings for always terminating rewrite derivations. This is 
important for confluence analysis.

\begin{note}[Independence from parameters]
No rule in \CANON or \SIMP contains any
parameter. This would not make sense anyway, as parameters are problem-specific
while the rules are meant to be problem-unspecific.
Observe that the orientedness test operates on rules only, without reference to any parameters.
This entails that Lemma~\ref{lemma:stability} and the claim of orientedness wrt.\ ordinary
instances made above applies in context of any set of parameters $\Pi$. As a consequence,
the ARI-normalization procedure can use $\AWPO(\epsilon)$ for
normalizing ground $\Sigma_\Pi$-terms is guaranteed to terminate.

The same remarks apply to the following confluence analysis. In brief, If a critical pair can
be shown joinable by our method, then all ground $\Sigma_\Pi$-instances will also be joinable.
\qed
\end{note}

\subsection{Local Confluence Analysis}
\label{sec:confluence-analysis}
A rewrite system $R$ is confluent if $R$ is terminating and locally confluent. This is
well-known. Given a terminating $R$, local confluence can be shown by analyzing 
\emph{critical-pairs}. A critical pair consists of two terms $s$ and $t$ obtained from
overlapping the left hand side of two rules in $R$ and represents potential case for
non-confluence. If $s$ and $t$ can be joined (both rewritten to some term $u$) then
confluence for $s$ and $t$ is established. If all critical pairs of all rules in $R$ are
confluent, then so is $R$. In the following $R$ is either \CANON or \SIMP as these are the
non-obvious cases.

In the following we sketch out method for critical pair analysis. It
generalizes standard  critical pair analysis by
taking ordering constraints and \kwote-terms into account. 
While a detailed treatment with full proofs is beyond the scope of this paper, we provide
high-level arguments for its correctness. 

\paragraph{Confluence analysis procedure.}
Let $\rho_1 = (l_1 \to r_1 \mid C_1)$ and $\rho_2 = (l_2 \to r_2 \mid C_2)$ be two rules, already made
variable disjoint by renaming. Our method starts with forming a critical \emph{triple} as follows.
Assume $l_1 = (l_1)_p[u]$, $u$ not a variable, and $u$ and $l_2$ have a most general unifier
$\sigma$. The critical triple has the three components $(l_1\sigma)_p[r_2\sigma]$ (``left'') , $r_1\sigma$
(``right''), and $C_1\sigma \cup C_2\sigma$ (``constr''). A critical triple represents the same situation as a critical
pair between ordinary rules $l_1 \to r_1$ and $l_2 \to r_2$ at the same position and same $\sigma$,
however restricted to ground instances that jointly satisfy $C_1$ and $C_2$.

The joinability test must prove that joinability holds for every ground instance of
the left and the right term, via some ground substitution $\gamma$, provided that 
the instance  $C_1\sigma\gamma \cup C_2\sigma\gamma$ is satisfied.
If $C_1\sigma \cup C_2\sigma$ unsatisfiable, the test succeeds trivially.
Satisfiability of (universally quantified) ordering constraints over $\succ$ can be tested via double
negation. For example, $\{ s_1 \succ t_1, s_2 \succ t_2\}$ is satisfiable iff neither $t_1 \succeq s_1$
nor $t_2 \succeq s_2$ holds, where $s \succeq t$ means ``$s \succ t$ or $s = t$ (syntactic equality).

If $C_1\sigma \cup C_2\sigma$ is satisfiable then the joinability test proceeds as follows.
Let $X$ be the set of variables in the critical triple. Every $x$ in $X$ represents an
unknown ground term $s$ (under further restrictions if $x$ is background).
The joinability test leaves the variables $X$ in place as Skolemized representations
of a set of unknown ground terms. If joinability is shown with the variables $X$ in
place, then ground joinability for ground instances straightforwardly follows.
We use normal-form computation $R\downarrow$ for that.
So far this is standard argumentation and has been used, e.g., for rewrite systems with domain constraints, see~\cite{kop_term_2013}, but things get more complicated now.

The AWPO for normalization is constructed and informed by the ordering constraints $C_1\sigma \cup C_2\sigma$ as follows.
Because $C_1\sigma \cup C_2\sigma$ in general has complex terms, the situation is more complicated than in the
termination test, where constraints are always between variables $x$ and $y$.

Suppose for now that $X$ has only two variables $x$ and $y$. They represent unknown ground terms, say, $s$
and $t$ respectively. Consider any AWPO $\succ$ that is ground-total ($\sem{0} \succ \sem{-1}$ for example), see Section~\ref{sec:ground-totality}. With that, exactly one of $s \succ t$, $s = t$ or
$t \succ s$ must hold. This observation lifts into a case analysis with corresponding assumptions
$x \succ y$, $x = y$ or $y  \succ x$.
For example, the $x \succ y$ assumption is exploited by taking the AWPO $\succ_{\AWPO(x > y)}$.
The procedure then tests satisfiability of $C_1\sigma \cup C_2\sigma$ wrt.\ $\succ_{\AWPO(x > y)}$. If satisfiable, normalization of
the left and right terms is tried with ordering $\succ_{\AWPO(x > y)}$. The $y \succ x$ case is symmetric. The
$x=y$ case is done by means of the substitution $\iota = \{x/y\}$: if $C_1\sigma\iota \cup C_2\sigma\iota$ is
satisfiable wrt.\ $\succ_{\AWPO(\varepsilon)}$ then the $\iota$-instances of left and right are tried for
normalization with ordering $\succ_{\AWPO(\varepsilon)}$.

If $X$ has more (or less) than two variables then the case analysis is generalized in the
obvious way. Every subset of $X$ has to be considered for making all its variables equal
with a substitution $\iota$, and for the remaining non-equal variables all possible order
arrangements have to be considered.

For our actual rewrite systems $R$ (\CANON or \SIMP) it turns out that further case distinction
is useful, to succeed more often. If $x$ is a foreground variable in the critical triple, an exhaustive case distinction can be made
whether $x$ represents a \kwote-term -- or not. In the first case $x$ is replaced by
$\sem{x}$, and in the second case $x$ is left in place but the additional precedence
constraint $x >
\kwote$ is added to the AWPO. This is possible because every non-\kwote term is greater
than every \kwote-term, in every AWPO.

Every case must succeed for the critical triple analysis to succeed as a whole.

Of course, in general these case distinctions lead to combinatorial explosion. 
In our applications the size of $X$ is at most 5 and was well manageable in terms of
resource consumption.

Equally crucial for success of confluence analysis is simplification.
Critical triples with \kwote-terms in left and right
can differ syntactically but still be $\calA$-equal. For the normalization with $R$ we
chose ARI-normal form computation as the simplification
function $\simp$. To simplify $t$ in a \kwote-term $\sem{t}$, hence, initially in left or right,
or derived by $R$, we compute $t \mathop{\to_\ARI} u$ for the underlying signature with
parameters $\Pi = X$ and obtain $\sem{u}$.
This is admissible because $\mathop{\to_\ARI}$ is sound and terminating; uniqueness of $u$
is preferable but not required (without it, joinable pairs might be missed).

\end{document}